\pdfoutput=1
\documentclass[11pt]{article}

\usepackage{styles/acl}

\usepackage{times}
\usepackage{latexsym}

\usepackage[T1]{fontenc}
\usepackage[utf8]{inputenc}

\usepackage{microtype}

\usepackage{todonotes}
\usepackage{subcaption}                     % for subfigures and subtables
\usepackage{graphicx}
\usepackage[nolist]{acronym}                % abbreviations
\usepackage{hyperref}                       % hyperlinks
\usepackage{xurl}                           % simple URL typesetting
\usepackage{booktabs}                       % professional-quality tables
\usepackage{multirow}                       % Row spanning multiple rows
\usepackage{amsfonts}                       % blackboard math symbols
\usepackage{nicefrac}                       % compact symbols for 1/2, etc.
\usepackage{enumitem}                       % custom spacing in enum, item envs
\usepackage{pgfplots}                       % plots
\pgfplotsset{compat=1.17} 
\usepackage{pgfplotstable}

\author{Joel Niklaus \and Daniele Giofr\'e  \\
  Thomson Reuters Labs, Zug, Switzerland \\
  \texttt{{firstname}.{lastname}@thomsonreuters.com}
}

\title{BudgetLongformer: Can we Cheaply Pretrain a SotA Legal Language Model From Scratch?}

\begin{document}
\maketitle

% Submit to https://neurips2022-enlsp.github.io/ and ARR

% Submit to IJCAI or ICAIL: https://icail2023.di.uminho.pt/

% After conversation with Valentin Hofmann: He told me that to get the best paper award, your paper has to be rejected at least once. Three times 2.5 on ARR is not that bad, just make minor changes and submit to ACL.

\begin{acronym}[UMLX]
    \acro{MLM}{Masked Language Modeling}
    \acro{RTD}{Replaced Token Detection}
    \acro{GAN}{Generative Adversarial Network}
    
    \acro{US}{United States}
    \acro{EU}{European Union}
    \acro{ECtHR}{European Court to Human Rights}

    \acro{NLP}{Natural Language Processing}
    \acro{ML}{Machine Learning}
    
    \acro{LM}{Language Model}
    \acro{SOTA}{state-of-the-art}
    
    \acro{BERT}{Bidirectional Encoder Representations from Transformers}
    
    \acro{TC}{Text Classification}
    \acro{QA}{Question Answering}
    \acro{NER}{Named Entity Recognition}
\end{acronym}

\begin{abstract}
% Pretrained transformers have achieved sota in many tasks
Pretrained transformer models have achieved state-of-the-art results in many tasks and benchmarks recently.
% SOTA LMs do not scale well above 512 tokens
Many state-of-the-art Language Models (LMs), however, do not scale well above the threshold of 512 input tokens.
% In the legal domain models often need to process very long text
In specialized domains though (such as legal, scientific or biomedical), models often need to process very long text (sometimes well above 10000 tokens).
% Although we have efficient models, they are often not available for special domains
Even though many efficient transformers have been proposed (such as Longformer, BigBird or FNet), so far, only very few such efficient models are available for specialized domains.
% Additionally, pretraining is very expensive and only in reach of large research labs
Additionally, since the pretraining process is extremely costly in general -- but even more so as the sequence length increases -- it is often only in reach of large research labs.
% ELECTRAs RTD task makes it cheaper
One way of making pretraining cheaper is the Replaced Token Detection (RTD) task, by providing more signal during training, since the loss can be computed over all tokens.
% We train a longformer with the RTD task on legal data
In this work, we train Longformer models with the efficient RTD task on legal data to showcase that pretraining efficient LMs is possible using much less compute. 
% We evaluate it on challenging downstream summarization tasks
We evaluate the trained models on challenging summarization tasks requiring the model to summarize long texts to show to what extent the models can achieve good performance on downstream tasks.
% We find that both small and base models are new sota on BillSum in their parameter range
We find that both the small and base models outperform their baselines on the in-domain BillSum and out-of-domain PubMed tasks in their respective parameter range.
% We open source our code and our models?
We publish our code and models for research purposes.
\end{abstract}

\acresetall % Reset acronyms for better readybility

% We are pushing the Pareto frontier, by presenting a good model using much less compute

\section{Introduction}

% Main points:
% - very little pretraining works well
% - Electra works well for Longformer
% - works very well for BillSum
% - given very little pretraining, we are very good already
% - LexGLUE is not our focus (because mostly short context), but we tested for completeness and we are ok, without tuning hyperparameters that are probably optimized for Legal-BERT

% Pretrained Transformer success blabla
Pretrained transformer models have achieved excellent performance across various \ac{NLP} tasks such as \ac{TC}, \ac{NER}, \ac{QA} and summarization \cite{devlin_bert_2019, yang_xlnet_2020, he_debertav3_2021, zhang_pegasus_2020}.

% Pretraining is expensive, 
Transfer learning is to a large extent responsible for this success \cite{howard_universal_2018}. Usually, transformer models are pretrained in a self-supervised way on large unlabeled corpora \cite{devlin_bert_2019, radford_improving_2018}. Pretraining is very resource intensive (especially for large models), thus making it costly and only available for large organizations \cite{sharir_cost_2020}. The \ac{MLM} task has been very successful, with many models adopting the task in their pretraining \cite{devlin_bert_2019, liu_roberta_2019, beltagy_longformer_2020, zaheer_big_2021}. Since typically only 15\% of the tokens are masked, the loss can be computed for those tokens only. 

% but ELECTRA can make it more affordable and more eco-friendly
\citet{clark_electra_2020} introduced the \ac{RTD} task, which enables the loss to be computed on all tokens, making training more efficient. On the GLUE benchmark \cite{wang_glue_2018}, their ELECTRA model matches RoBERTa \cite{liu_roberta_2019} and XLNet \cite{yang_xlnet_2020} using 1/4 their compute. 
% ELECTRA is a very convincing approach, but barely used in the literature
Although ELECTRA's training strategy seems very promising, to the best of our knowledge, only few works have adopted the \ac{RTD} task so far \cite{he_debertav3_2021, kanakarajan_bioelectrapretrained_2021}.

\begin{figure}[t]
    \centering
    \resizebox{\columnwidth}{!}{
        \includegraphics{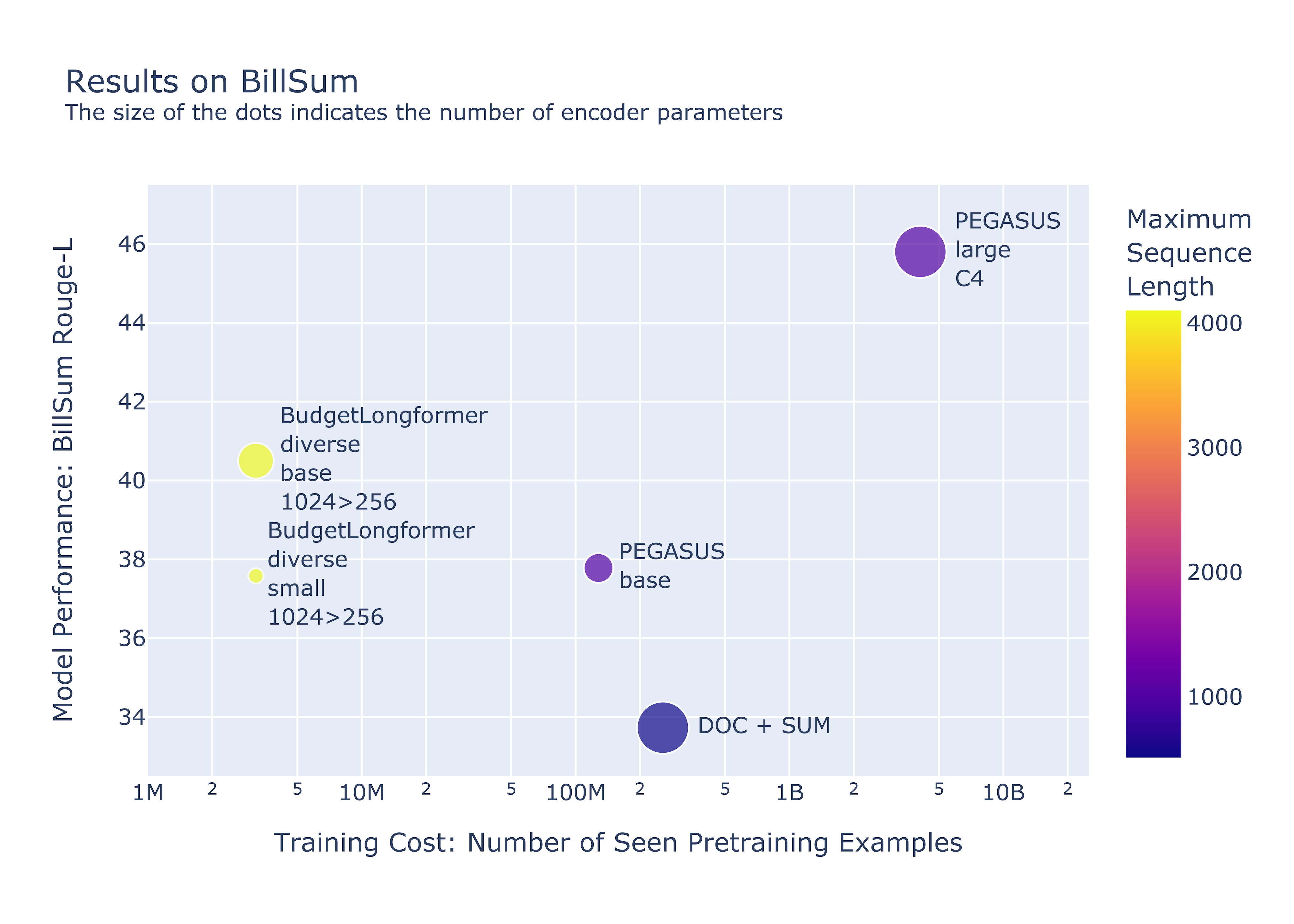}
    }
    \vspace{-7mm}
    \caption{Results on the BillSum dataset. Note that the x-axis is in log-scale.}
    \label{fig:billsum}
    \vspace{-5mm}
\end{figure}

% Domain Specific Models can provide boost
On another note, domain-specific pretraining has been shown to improve downstream performance in many domains such as 
law \cite{chalkidis_legal-bert_2020, xiao_lawformer_2021}, 
biology \cite{lee_biobert_2019},
scientific articles \cite{beltagy_scibert_2019},
clinical documents \cite{li_clinical-longformer_2022}, or even
code \cite{chen_evaluating_2021}. Domain-specific pretraining coupled with the \ac{RTD} task, however, has not been studied in the legal domain so far.

% Long document processing is important, especially in the legal domain
Depending on the domain, documents might be extremely long. Texts from the legal domain, for example, tend to span multiple pages, ranging from 10s to 100s of pages, which translates to tens of thousands tokens. The quadratic time and memory requirement of the attention typically used in the transformer architecture \cite{vaswani_attention_2017} prohibits efficient processing of sequences longer than 512 tokens on current hardware. A rich body of research investigates how transformers can be adapted to efficiently process longer input \cite{tay_efficient_2020, child_generating_2019, beltagy_longformer_2020, zaheer_big_2021, roy_efficient_2021, kitaev_reformer_2020, tay_synthesizer_2021, lee-thorp_fnet_2021}.
%Especially in the legal domain, texts tend to be extremely long, with documents exceeding 100 pages (approx. 50000 tokens). 

% So far, there does not exist a longformer model pretrained on English legal data
Longformer \cite{beltagy_longformer_2020} is one of these efficient transformer architectures for long sequences, leveraging windowed and global attention. So far, to the best of our knowledge, there does not yet exist a public Longformer model pretrained on English legal data\footnote{On the web there is a model based on Longformer in a legal domain but no link how it was obtained and on its actual performance (https://huggingface.co/saibo/legal-longformer-base-4096).}, although \citet{xiao_lawformer_2021} have proven the effectiveness of the Longformer in dealing with long legal text in many Chinese-related tasks. This work aims to fill this gap.

% Why we evaluate on Summarization
%To consider a task that could be used to assess the ability to grasp long-distance dependencies in the text, we evaluated our LMs mainly through the challenging task of automatic summarization. It consists of capturing the most important concepts/ideas from the (long) document and automatic rewriting it in a shorter passage while preserving a grammatically and logically coherent way \cite{chen_multi-task_2019}. 
To test the ability to grasp long-distance dependencies in the text, we mainly evaluated our \acp{LM} on the task of automatic (abstractive) summarization. It consists of capturing the most important concepts/ideas from the (long) document and then rewriting it in a shorter passage in a grammatical and logically coherent way \cite{chen_multi-task_2019}.

% BillSum and PubMed in particular
In particular, we used the BillSum benchmark, as a domain-specific summarization task, obtaining a new \ac{SOTA} (see Figure \ref{fig:billsum}); and the PubMed benchmark, to evaluate the model's ability outside the legal context (i.e., in the biomedical context), obtaining comparable metrics even though the \ac{LM} has only been pretrained on legal data and the tokenizer is also optimized for legal data (see Figure \ref{fig:pubmed}).

%Based on the previous statements, 
We emphasize that this performance was achieved with a minimal pretraining phase due to the combination of the RTD task and the Longformer infrastructure, %compared to other more resource-greedy architectures
making our \ac{LM} very attractive from the point of view of building costs. For instance, our model saw only 3.2M examples during pretraining, whereas RoBERTa \cite{liu_roberta_2019} or PEGASUS-large \cite{zhang_pegasus_2020} saw  4.1B examples. RoBERTa was trained for 1024 GPU days, whereas our small and base models only used 12 and 24 GPU days respectively (16GB NVIDIA V100 GPUs for both models).

% HierBERTs can only handle text classification out of the box
Since many tasks in legal \ac{NLP} are formulated as \ac{TC} problems, a hierarchical architecture has been used frequently to process long documents \cite{chalkidis_neural_2019, niklaus_swiss-judgment-prediction_2021}. This simple hierarchical architecture, indeed, cannot be easily adapted to solve the more complex sequence-to-sequence tasks like token classification or summarization, because it do not take efficiently long input correlations. For this reason, in this work, we pretrain a more versatile Longformer model.

% LexGLUE for completeness
Finally, for completeness, we evaluated our \acp{LM} using the LexGLUE benchmark, which is mainly based on multi-class and multi-label legal \ac{TC} problems for short texts. 

% Our model reaches very good performance using very few parameters and very little pretraining compute
% We show that, on a challenging summarization dataset, our model outperforms a model trained for ten times longer and containing almost 12 times more encoder parameters.
% \todo{add more here once we have more results}

\subsection*{Contributions}
The contributions of this paper are five-fold:
\vspace{-4mm}
\begin{itemize}[leftmargin=8pt, itemsep=0em]
    % release long legal model
    \item We train and release  %, for the first time, 
     a new model pretrained on recently published curated English legal text  \cite{henderson_pile_2022}, capable of handling input spans longer than 512 tokens out of the box.
    % Apply electra task to longformer
    \item We apply the promising, but seldom used \ac{RTD} task \cite{clark_electra_2020} on a Longformer model \cite{beltagy_longformer_2020}, for the first time, calling it BudgetLongformer.
    % Outperform larger models on LexGLUE
    % Outperform larger models on BillSum
    \item On the BillSum benchmark \cite{kornilova_billsum_2019}, our models are a new \ac{SOTA} compared to models of the same size. Especially, our small model outperforms all baseline approaches, and a transformer base model \cite{vaswani_attention_2017} containing almost 4 times more encoder parameters (110M vs. 29M). It even outperforms the PEGASUS base model \cite{zhang_pegasus_2020} whose encoder is also almost 4 times larger and has been pretrained specifically for the abstractive summarization task in mind.
    \item We verified that pretraining with the RTD task is suitable for down-stream summarization tasks by evaluating our model on an out-of-domain benchmark (PubMed), obtaining comparable results with summarization-specific architectures.
    \item  On the LexGLUE benchmark \cite{chalkidis_lexglue_2021}, despite the obvious emphasis on covering classification tasks even for short documents, our models achieve metrics equivalent to those of architectures that are better suited to this length of document, and with substantially fewer numbers of parameters and pretraining steps.
    %our LMs outperforms \ac{SOTA} distilled models such as miniLM \cite{wang-etal-2021-minilmv2} or DistilBERT \cite{sanh_distilbert_2020} using 4 times less compute (100K steps vs 400K distillation steps for miniLM excluding training of the vanilla model) and more than 2 times fewer parameters (29M vs. 66M). Our method is a great way of cheaply developing a model when no large vanilla model is available. \todo{Maybe mention this in the introduction as well}
    % Stress that we enable long-context tasks like summarization or question answering that cannot be solved with hierarchical transformer setups
\end{itemize}

%\todo{Read ELECTRA paper again}

%\todo{Read papers that also make the point of being less costly or having less parameters}

% This can easily be commented out in case we have too much content.
%In the remainder of this paper, we discuss prior work in Section \ref{sec:related_work}. We describe the datasets we used in Section \ref{sec:datasets}, the experimental setup in Section \ref{sec:experimental_setup} and the results in Section \ref{sec:results}. Finally, we answer the main research questions, discuss limitations, draw conclusions and give directions for future work in Section \ref{sec:conclusions_future_work}.

\subsection*{Main Research Questions}
In this work, we pose and examine five main research questions:

% Hypothesis: RTD task could be even better on legal data than on normal data than MLM task (may be reason for why we overperform on BillSum). News language is less strict on vocabulary than legal language

\noindent \textbf{RQ1}: \emph{Is it possible to generate an ad-hoc \ac{LM} with domain (e.g. legal) expertise from scratch, reducing costs and CO$_2$ emissions?}

\noindent \textbf{RQ2}: \emph{Is it possible to pretrain a Longformer model with the \ac{RTD} task (aka BudgetLongformer)?}

\noindent \textbf{RQ3}: \emph{How does our BudgetLongformer compare with other models on the challenging summarization task? Particularly in the case of a legal domain-specific benchmark such as BillSum?}

\noindent \textbf{RQ4}: \emph{ How well does our BudgetLongformer generalize to other domains, for example in the biomedical domain, as evaluated by the PubMed summarization benchmark?}

\noindent \textbf{RQ5}: \emph{How do our \acp{LM} compare with other models on the \acf{TC} benchmark LexGLUE?}

\section{Related Work}
\label{sec:related_work}

\subsection*{Domain-Specific Language Models}
% This is directly copied from the Intro. Better to rephrase or exclude
% Legal BERT, SciBERT, BioBERT, etc.
Previous work showed that domain-specific pretraining shows promising results on datasets of specialized domains such as 
law \cite{chalkidis_legal-bert_2020, xiao_lawformer_2021}, 
biology \cite{lee_biobert_2019},
scientific articles \cite{beltagy_scibert_2019},
clinical documents \cite{li_clinical-longformer_2022}, or even
code \cite{chen_evaluating_2021}. % code

% don't stop pretraining: domain adaptation
\citet{gururangan_dont_2020} show that continued pretraining on a RoBERTa checkpoint on biomedical data, scientific articles in computer science, and reviews, clearly improves downstream performance in the respective domain-specific datasets. The effect was less pronounced on datasets from the news domain, presumably because RoBERTa has seen many news articles in its pretraining already.

\subsection*{Long Document Processing}
% Longformer, BigBird, FNet, Long Range Arena, Tay Survey, etc.

% copied from research proposal
In the past few years, a vast amount of research has been devoted to addressing the problem of quadratic time and memory complexity associated with the dense attention mechanism \cite{vaswani_attention_2017}, practically limiting the maximum sequence length severely (often to 512 tokens) \cite{tay_efficient_2020, child_generating_2019, beltagy_longformer_2020, zaheer_big_2021, roy_efficient_2021, kitaev_reformer_2020, tay_synthesizer_2021, lee-thorp_fnet_2021}. These research works have given rise to a new class of transformers, referred to as sparse transformers or efficient transformers \cite{tay_efficient_2020}. Reducing the cost associated with the computation of the dense attention matrix while maintaining the same performance is the core idea behind efficient transformers. This is often achieved by introducing sparsity in the attention matrix in a variety of ways that may be fixed pattern such as local (windowed) attention \cite{child_generating_2019, beltagy_longformer_2020}, global attention \cite{zaheer_big_2021} or learnable patterns such as routing attention \cite{roy_efficient_2021} and LSH attention \cite{kitaev_reformer_2020} or a random pattern \cite{zaheer_big_2021, tay_synthesizer_2021}. Recently, \citet{lee-thorp_fnet_2021} proposed to use Fourier transforms instead of the attention layer. A comprehensive list of efficient transformers and the detailed description of their attention mechanism can be found in the survey by \citet{tay_efficient_2020}.
\cite{tay_long_2020} proposed a series of tasks designed for testing the capabilities of these different models suitable for longer inputs. However, this so-called ``Long Range Arena'' considers mostly artificial tasks, with the goal of evaluating the models independently of any pretraining.

\subsection*{Efficient Pretraining}
% ELECTRA
ELECTRA-style pretraining \cite{clark_electra_2020} has been shown to reduce training cost substantially, while matching the performance of \ac{SOTA} \acp{LM}. ELECTRA leverages a smaller generator model (discarded after pretraining), that changes some tokens. The larger discriminator model (used for down-stream tasks) must predict for each token if it was changed by the generator or not, similar to how \acp{GAN} are trained \cite{goodfellow_generative_2014}. This enables the loss to be relevant for every token, leading to much faster and thus more efficient training.

\section{Datasets}
\label{sec:datasets}
In this section, we briefly introduce the datasets used in our experiments.

\subsection{Pile of Law}
\citet{henderson_pile_2022} recently released a large-scale English corpus suitable for pretraining \acp{LM}. It contains 256 GB of diverse legal text in English from various jurisdictions and judicial bodies including for example bills, court decisions and contracts from the US, Canada, and Europe even though the focus clearly lies on US data. 
While there are 28 US datasets available (253.25 GB or 99\%), there is only 1 Canadian dataset\footnote{Canadian Court Opinions (ON, BC)} (243 MB or 0.09\%), 3 European datasets\footnote{European Court of Human Rights Opinions, EUR-LEX and European Parliament Proceedings Parallel Corpus} (2.3 GB or 0.9\%), and 2 international datasets\footnote{World Constitutions and U.N. General Debate Corpus} (212 MB or 0.08\%). 
The non-US datasets only cover the categories ``Legal Case Opinions and Filings'', ``Laws'' and ``Conversations'', but do not cover categories ``Legal Analyses'', ``Contracts / Business Documents'' and ``Study Materials'', whereas the US data is much more diverse and covers all categories.

\subsection{BillSum}

\citet{kornilova_billsum_2019} introduced a legislative summarization dataset from 21K US bills from 1993 to 2018. It is challenging due to the technical nature and complex structure of the bills. Additionally, the bills are rather long, ranging from 5K to 20K characters ($\sim$ 1K to 4K tokens\footnote{Our experiments show that using our tokenizer one token corresponds to 5.33 characters on average.}) with their summaries being up to 5K characters ($\sim$ 1K tokens) long (see Appendix \ref{sec:data_details} for more details). 

\subsection{PubMed}
\citet{cohan_discourse-aware_2018} introduced another challenging summarization dataset in a specialized domain (scientific articles from the biomedical domain). It includes 133K scientific papers together with their abstracts in English. The papers are 3K words long on average and the summaries (abstracts) 200 words. Thus, similar to the BillSum dataset, this dataset is well suited as a test bed for methods capable of long document summarization. Note, that in this dataset the domain is vastly different from the legal domain (see Appendix \ref{sec:data_details} for more details).

\subsection{LexGLUE}
\citet{chalkidis_lexglue_2021} recently introduced a benchmark for the English legal domain called LexGLUE. LexGLUE contains six \ac{TC} tasks and one \ac{QA} task comprising diverse legal data such as US court decisions and contracts, terms of service documents, EU legislation and cases from the \ac{ECtHR}. There exists a public leaderboard of diverse models on GitHub\footnote{\url{https://github.com/coastalcph/lex-glue}}, with Legal-BERT \cite{chalkidis_legal-bert_2020} performing best. 

The LexGLUE benchmark focuses on evaluating \acp{LM} in legal \ac{TC} and \ac{QA} tasks. 
%To make the procedure fair to short-span models (e.g. BERT, RoBERTa, Legal-BERT), LexGLUE has many tasks that require less than 512 tokens.
In LexGLUE, 4 out of 7 tasks involve documents with input lengths lower than 512 tokens on average. From the remaining 3 tasks, the ECtHR A and B tasks and the SCOTUS tasks involve documents with long span, and  the median of the first two is also less than 1000 tokens. Usually, legal documents are much longer than 512 tokens and thus this distribution might not be representative of real-world tasks. Shorter input length tasks may be better handled by short-input models (e.g., BERT, RoBERTa, Legal-BERT, etc.).

\section{BudgetLongformer}
% we trained a domain specific longformer model with the electra task
In the legal domain, it is especially important that models can handle long input. So far, there does not exist an English legal model capable of handling more than 512 tokens. To make pretraining more affordable, we combined the well-proven Longformer model \cite{beltagy_longformer_2020} with the \ac{RTD} task proposed by \citet{clark_electra_2020}.

%\todo{Write more about this: Introduce how electra and longformer work and what we do. We use Longformer global + sliding window version because we train an encoder only model}

%\todo{In footnote, mention that we could imagine several ablation studies (1. pretrain longformer with MLM on our data and evaluate, 2. pretrain our method on other data and evaluate, 3. investigate different tokenizer sizes and ablate) which we leave for future work because of missing computational resources (we already spent 144 (pretraining) + 12 (debugging) + 25 (BillSum) + 7 (PubMed) + 11 (LexGLUE) = 199! GPU days for our experiments) ==> costs (~3USD per GPU hour only for GPU + additional costs for storage, transfer, etc.) ==> 24 * 3 * 200 = 14400USD}

%\todo{Mention that we trained the models for up to 200K steps, but did not see an improvement in downstream performance }

\section{Experimental Setup}
\label{sec:experimental_setup}
In this section, we describe how we set up the experiments.
In all our experiments, we made use of AMP mixed precision training and evaluation to reduce costs and GPU memory. For all our experiments, we used the huggingface transformers library \cite{wolf_transformers_2020} available under an Apache 2.0 license.

\subsection{Tokenizer}
We trained a byte-level BPE tokenizer \cite{wang_neural_2019} similar to \citet{beltagy_longformer_2020}. To encode the complicated legal language well, we chose a relatively large vocabulary of 64K tokens (additionally, we did not apply any preprocessing/cleaning of the input texts). We trained the tokenizer using the huggingface tokenizers library\footnote{\url{https://github.com/huggingface/tokenizers}} on the entire PileOfLaw training split ($\sim$ 192GB, $\sim$ 22.5B tokens, $\sim$ 7.5M documents), covering a wide array of English legal texts, mostly from the US.

\subsection{Pretraining}

\begin{table}[t]
\centering
\resizebox{\columnwidth}{!}{
    \begin{tabular}{lrrr}
    \toprule
    PileOfLaw Subset                    & Dataset Size & \# Words   & \# Documents \\ 
    \midrule
    caselaw \\
    \midrule
    \bf CL Opinions                     & \bf 59.29GB  &\bf 7.65B & \bf 3.39M\\
    \midrule
    diverse \\
    \midrule
    \bf Total                           & \bf 73.04GB  &\bf 8.91B & \bf 2.1M\\
    CL Opinions                         & 8.74GB  &  1.13B & 500K\\
    CL Docket Entries and Court Filings & 17.49GB &  1.80B & 500K\\
    U.S. State Codes                    & 6.77GB  &  829.62M & 157\\  
    U.S. Code                           & 0.27GB &  30.54M & 43\\    
    EUR-Lex                             & 1.31GB  &  191.65M & 106K\\    
    Edgar Contracts                     & 7.26GB  &  0.97B & 500K\\    
    Atticus Contracts                   & 31.2GB  &  3.96B & 488K\\    
    \bottomrule
    \end{tabular}
}
\caption{The datasets used for pretraining our models. CL is short for Court Listener}
\label{tab:pretraining_data}
\end{table}

We trained the \emph{caselaw} models on the training subset ``Court Listener Opinions'' from the PileOfLaw (59.3 GB, 7.65B words, 3.39M documents).
The \emph{diverse} models were trained on caselaw (``Court Listener Opinions'' \& ``Court Listener Docket Entry Documents''), legislation (``US Code'', ``State Codes'' \& ``EURLEX'') and contracts (``Atticus Contracts'' \& ``EDGAR Contracts''). 
To balance the training data, we limited the number of documents to 500K (this affects Court Listener Opinions, Court Listener Docket Entry Documents and EDGAR Contracts. Please see Table \ref{tab:pretraining_data} for more details.
Our validation set consisted of 1000 randomly selected examples from the respective training set.\footnote{We used such a relatively small validation set to save compute.}

To maximally use the available data, we concatenated all the examples and then cut them off in slices of the model's maximum sequence length (4096). We did this in batches of 1000 examples with multiprocessing to speed up data preparation. The last slice in each batch will not contain 4096 tokens, so we dropped it.

We trained both a small (29M parameters) and a base (159M parameters) model for each configuration. To reach 100K steps it took a bit less than 3 days for the small model and a bit less than 6 days for the base model on 4 16GB NVIDIA V100 GPUs. The achieved training and evaluation losses are shown in Table \ref{tab:losses} in Appendix \ref{sec:pretraining_details}. Interestingly, we find that the diverse models achieve lower training and evaluation losses. Please find more details in Appendix \ref{sec:hyperparameters}.

% \todo{Daniele: Rewrite}
% Diverse data can lead to unstable training 
\citet{henderson_pile_2022} have experienced difficulties when the language model was trained on the entire Pile-of-Law. We believe that the highly imbalanced dataset concerning text types (contracts, court decisions, legislation, etc.) is the main reason for the training instability. This is one of the reasons why we adopted the procedure described above. As shown later in the results (see Section \ref{sec:results}), our pretraining was stable. On the contrary, the diverse model -- includes more lexical and layout diversity of documents -- turns out to perform better and train more robustly on the summarization tasks.
%\todo{Revise last sentence}

%experienced training instabilities when training their BERT-large model on the entire PileOfLaw dataset, so they reduced the learning rate to 5e-6 and the batch size to 128, yielding stable training. 
%They only evaluate on the CaseHOLD dataset \cite{zheng_when_2021} and do not outperform a model trained exclusively on caselaw (CaseLawBERT \cite{zheng_when_2021}). 
% From PileOfLaw Paper: This is consistent with a number of recent results that suggest that masked-language model pretraining efficacy may saturate [https://arxiv.org/abs/2205.06266], especially as more diverse or distinct data sources are added [https://arxiv.org/abs/2110.02095], or may only give significant gains for highly in-domain data \cite{zheng_when_2021}.

% But in our case, when we carefully select the data sources the diverse models train well

\subsection{Downstream Benchmarks}

\subsubsection*{BillSum}
When finetuning on the BillSum dataset \cite{kornilova_billsum_2019} we trained using early stopping with patience of 3 epochs. We paired our pretrained encoder model with a randomly initialized bart-base decoder model \cite{lewis_bart_2020}.\footnote{Interestingly, the randomly initialized decoder yielded better results than when we used the weights from the pretrained huggingface checkpoint at \url{https://huggingface.co/facebook/bart-base}.} We used a batch size of 32 and learning rate of 7e-5 after tuning in \{5e-4, 9e-5, 7e-5, 5e-5, 3e-5, 1e-5\}. We used the bart-base default config for num\_beams (4) and no\_repeat\_ngram\_size (3). 
We set the maximum input length to 1024 and the maximum target length to 256 to save compute. However, many summaries get cut off at 256 tokens. This is why we took our best model and trained it with maximum input length 4096 and maximum target length 1024 (see results in Table \ref{tab:billsum_results} and examples in Table \ref{tab:examples_billsum-4096-1024_base_diverse}).
Due to high training costs, we only trained it with one random seed (42).
Our models contain 29M (small) and 159M (base) parameters in the encoder and 96M parameters in the decoder resulting in a total of 125M (small) and 255M (base) parameters.

\subsubsection*{PubMed}
Additionally, we evaluated on the PubMed summarization task \cite{cohan_discourse-aware_2018} using the same settings as for the BillSum task. We set the maximum input length to 4096 and the maximum generation length to 512.

\subsubsection*{LexGLUE}
Finally, we evaluated on LexGLUE \cite{chalkidis_lexglue_2021} using the publicly available scripts without modification to ensure consistent and comparable results. Because of compute limitations, we ran each experiment with only one random seed (1) and with the default set of hyperparameters. We speculate that hyperparameter tuning could further improve the performance of the proposed model. 
% We ran all experiments with 3 random seeds (1,2,3) and report the average and standard deviation.

%\todo{Possible further experiments if the paper quality is not good enough:
%- evaluate on cuad: https://github.com/huggingface/transformers/tree/main/examples/pytorch/question-answering
%- do more tuning on the MultiLexSum dataset
%- Evaluate on one of these tasks additionally: %https://github.com/neelguha/legal-ml-datasets#summarization
%- Evaluate on out of domain test set (ca_test) from BillSum
%- Evaluate sources to long task on MultiLexSum (not only long -> short, long -> tiny, short -> tiny)
%- Evaluate on big_patent (or at least a part of it since it is very large)
%- Evaluate on arxiv summarization task
%- Tune hyperparams on LexGLUE to get better results
%}

\section{Results}
\label{sec:results}

In the following three sections, we present the results on the BillSum dataset, the PubMed dataset and the LexGLUE benchmark. 
Tables \ref{tab:summarization_models} and \ref{tab:lexglue_models} in Appendix \ref{sec:models_overview} compare the models evaluated on the summarization and LexGLUE benchmarks, respectively.

% Our models use much fewer pretraining steps

% Not only pretraining steps are relevant but also batch size

% Number of seen examples combines these two metrics (pretraining steps and batch size)

\subsection{BillSum}

Our results on the BillSum dataset are presented in Figure \ref{fig:billsum} and Table \ref{tab:billsum_results} in Appendix \ref{sec:detailed_results}.

% small model
We observe that even our small diverse model
clearly exceeds the baseline of the original article (DOC + SUM), even though their model is based on BERT-large which contains almost 12 times more encoder parameters and has been pretrained for 10 times more steps. 
%Additionally, we also outperform a Transformer-base model containing almost twice as many parameters as our model.
Even more surprisingly, our small diverse model is on par with the PEGASUS-base model \cite{zhang_pegasus_2020} (37.58 vs. 37.78 Rouge-L), pretrained using the Gap-Sentences task specifically designed for abstractive summarization. Furthermore, their model contains almost 4 times more encoder parameters and has seen 40 times more training examples during pretraining (128M vs. 3.2M; see Table \ref{tab:summarization_models} in Appendix \ref{sec:models_overview}). 

% base model
By scaling up our model to the base size, we even approach the performance of PEGASUS-large (40.5 vs. 45.8 Rouge-L). PEGASUS-large has seen three orders of magnitude more training examples during its pretraining in comparison to our model (4.1B vs. 3.2M) and contains more than twice as many encoder parameters (340M vs. 159M).

We conclude that pretraining with the \ac{RTD} task is highly effective, with minimal compute for long-input summarization in-domain.

%We hypothesize that the domain specific pretraining and tokenizer are responsible for these improvements. Even though the vocabulary size of the PEGASUS models is larger (96K vs 64K), we suspect that the domain specific tokenizer is able to encode the input text more efficiently than the PEGASUS tokenizer. ==> No, the tokenizer of PEGASUS uses less tokens on average than ours on the BillSum training set

\subsection{PubMed}

\begin{figure}[t]
    \centering
    \resizebox{\columnwidth}{!}{
        \includegraphics{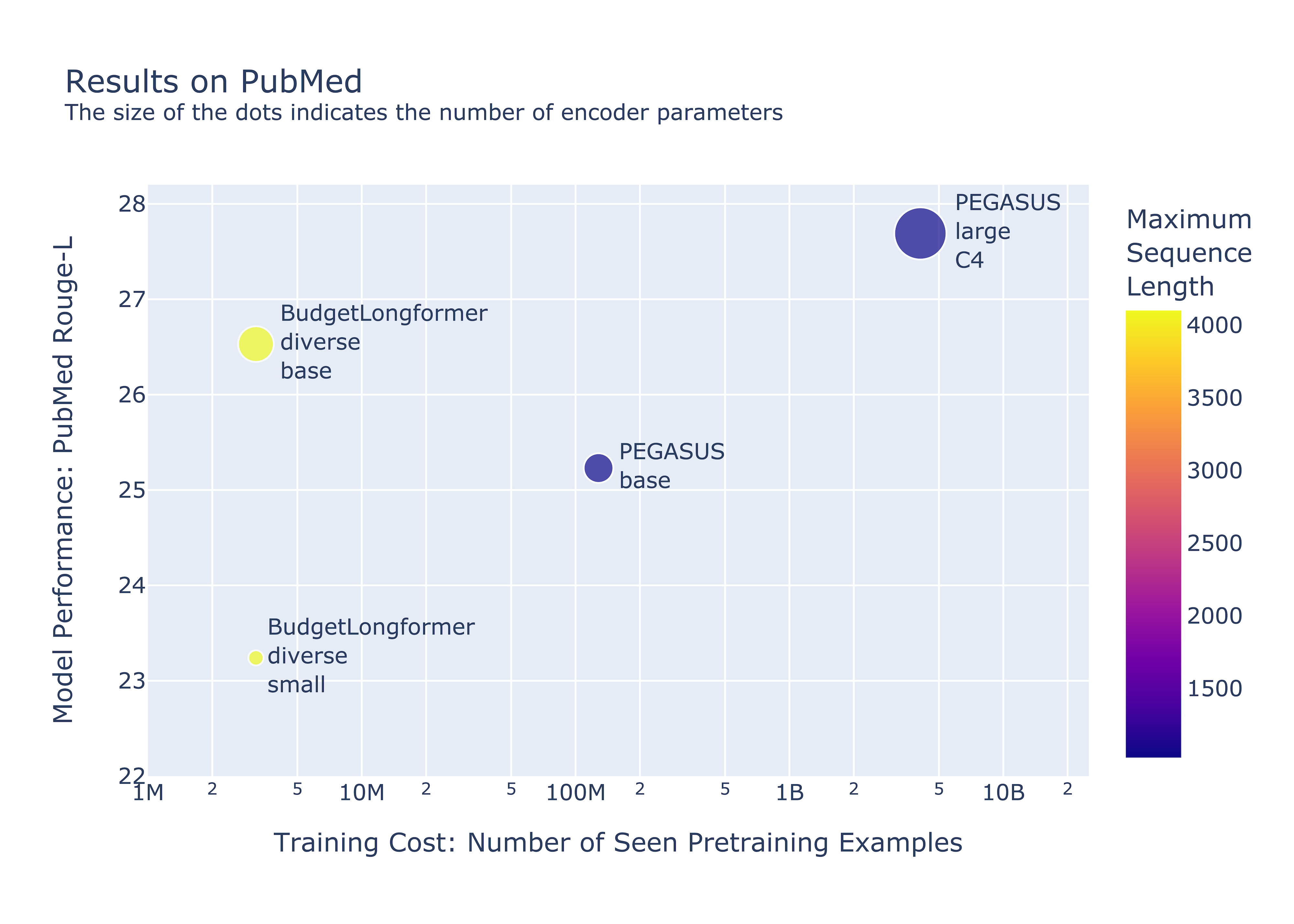}
    }
    \vspace{-7mm}
    \caption{Results on the PubMed dataset. Note that the x-axis is in log-scale.}
    \label{fig:pubmed}
    \vspace{-5mm}
\end{figure}

Our results on the PubMed dataset are presented in Figure \ref{fig:pubmed} and Table \ref{tab:pubmed_results}  in Appendix \ref{sec:detailed_results}.

% we outperform PEGASUS-base (SOTA from 2 years ago)
Similar to the results on BillSum, our small model clearly outperforms the Transformer-base model (23.24 vs. 19.02 Rouge-L) and approaches the PEGASUS-base model (23.24 vs. 25.2 Rouge-L)  even though we did not specifically pretrain our model for summarization and our model has seen 40 times fewer examples during pretraining (3.2M vs. 128M). 
%Note that even our small model outperforms the PEGASUS-base model (26.03 vs 25.23 Rouge-L).
Similar again, we almost reach the performance of PEGASUS-large (26.53 vs. 27.69 Rouge-L) while having seen 1280 times fewer examples during pretraining (3.2M vs. 4.1B).

% we did not train on medical data at all, even the tokenizer did not see it
Note, that we pretrain on a much narrower domain than PEGASUS (legal text vs. C4). Our tokenizer and model has never seen medical data during its pretraining phase.
% Our vocabulary size is smaller than that of PEGASUS
Finally, our tokenizer has $1/3$ fewer tokens than the PEGASUS tokenizer (64K vs. 96K).  

In conclusion, pretraining with the \ac{RTD} task is even effective on an out-of-domain downstream summarization task.

%\todo{Do Human Evaluations?}

\subsection{LexGLUE}

%\todo{Put everything LexGLUE related into the appendix.}

\begin{figure}[t]
    \centering
    \resizebox{\columnwidth}{!}{
        \includegraphics{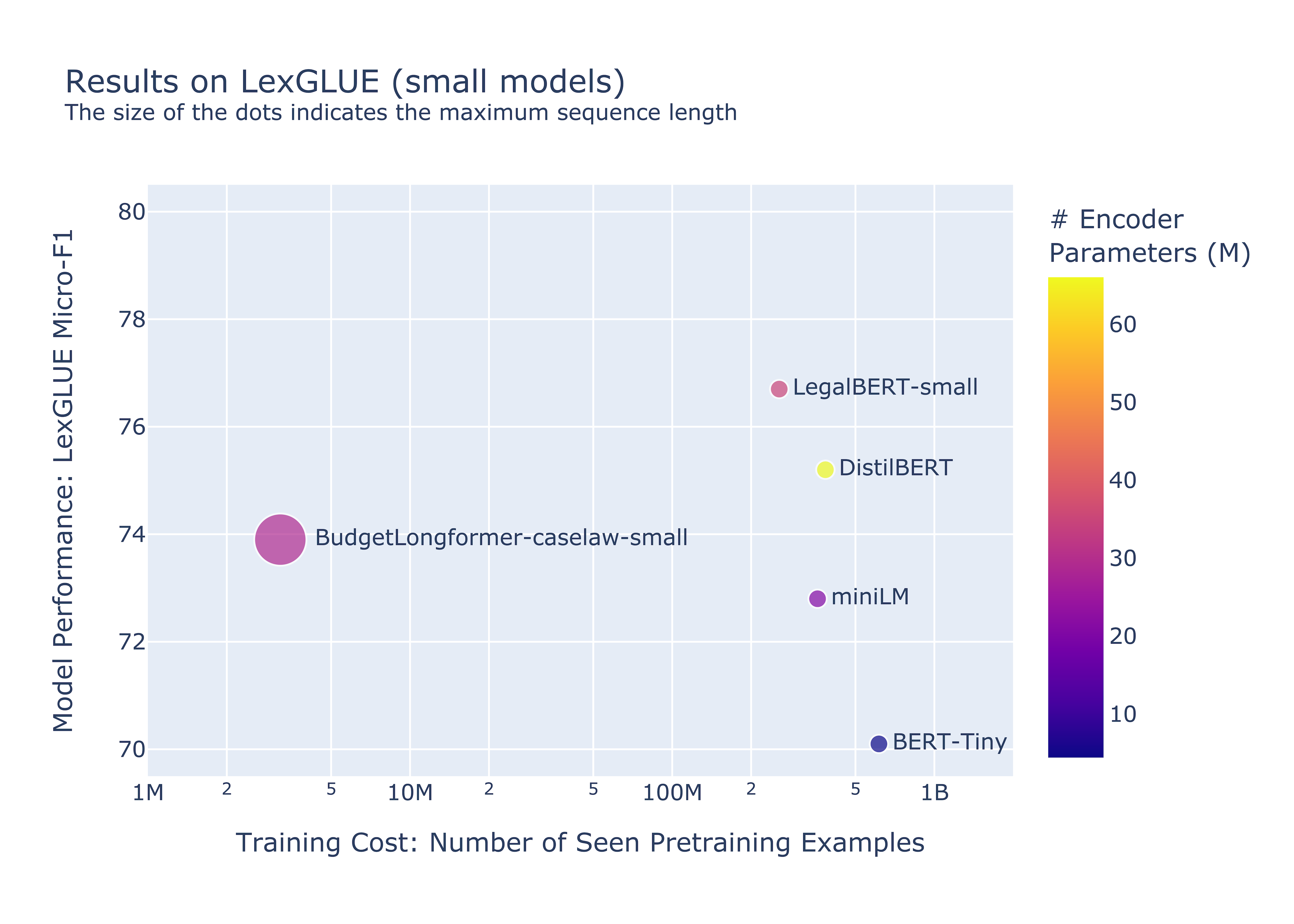}
    }
    \vspace{-7mm}
    \caption{Results on the LexGLUE benchmark (small models). Note that the x-axis is in log-scale.}
    \label{fig:lexglue_small}
    \vspace{-5mm}
\end{figure}

\begin{figure}[t]
    \centering
    \resizebox{\columnwidth}{!}{
        \includegraphics{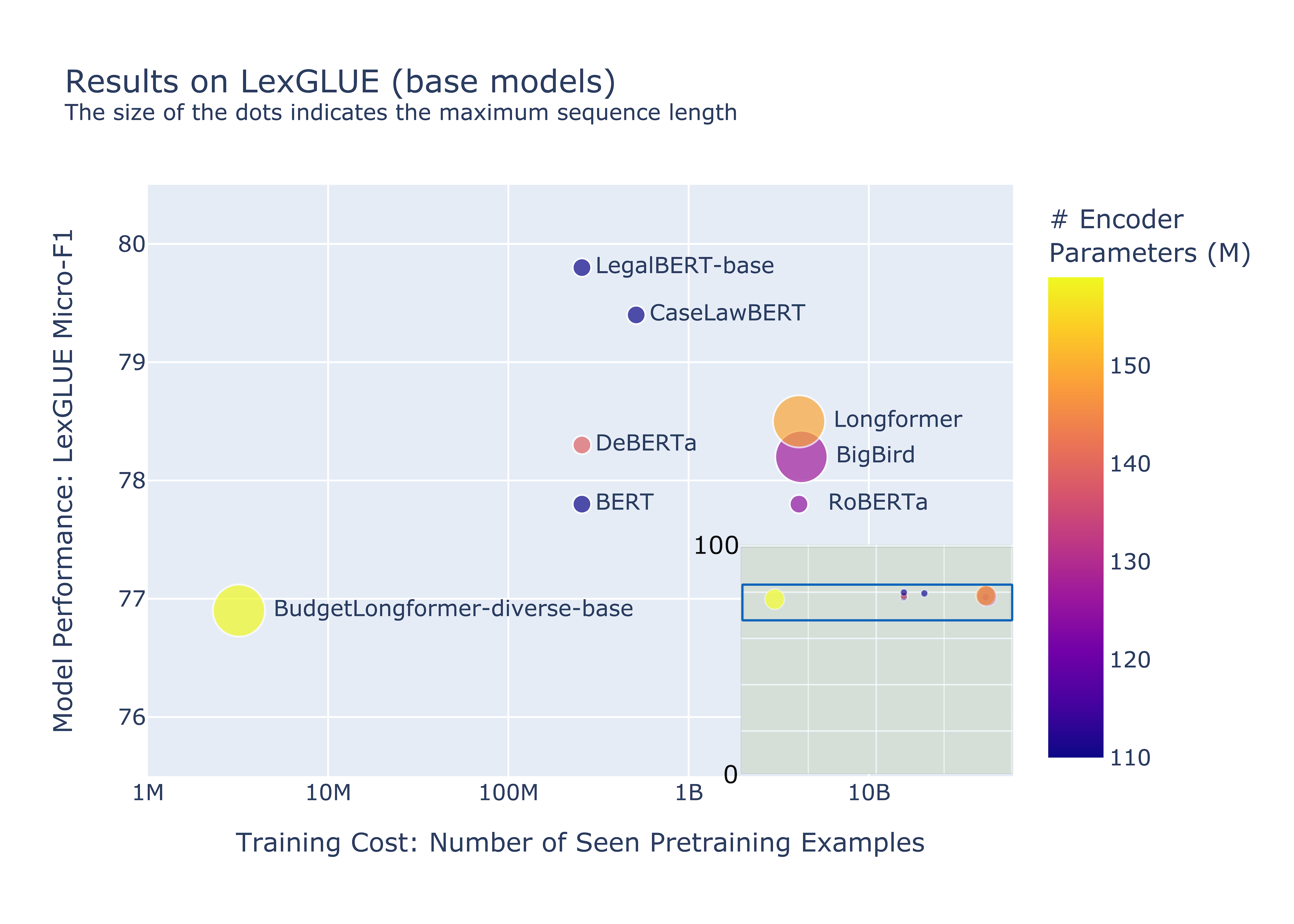}
    }
    \vspace{-7mm}
    \caption{Results on the LexGLUE benchmark (base models). Note that the x-axis is in log-scale.}
    \label{fig:lexglue_base}
    \vspace{-5mm}
\end{figure}

% We did not tune any hyperparameters but just took the default ones! Tuning might yield quite different results

Table \ref{tab:lexglue_models}  in Appendix \ref{sec:models_overview} compares the models evaluated on the LexGLUE benchmark. 
Note, that these models differ strongly on many dimensions such as the number and types of training steps, the architecture, and the number of parameters. 
%MiniLM, BERT-Tiny and DistilBERT were all distilled from BERT-base (1M steps). MiniLM was distilled for 400K steps, but for DistilBERT we were not able to find the number of pretraining steps, so we assumed that it was distilled for 500K steps. Our model was pretrained for 100K \ac{RTD} steps. Regarding number of parameters, the models range from 4.4M (BERT-Tiny) to 66M (DistilBERT).

Our results on the LexGLUE benchmark are presented in Table \ref{tab:lexglue_results} in Appendix \ref{sec:detailed_results} and in Figures \ref{fig:lexglue_small} and \ref{fig:lexglue_base} for the small and base models respectively. Figure \ref{fig:lexglue} in Appendix \ref{sec:detailed_results} shows all the models evaluated on LexGLUE combined.

% We are mainly bad in short-context tasks
From the results shown in Table \ref{tab:lexglue_results}, we can observe that our models do not improve on the \ac{SOTA} for short input length tasks. This suggests that for such tasks a more accurate description of the first 512 tokens, obtained through a pretraining dataset with a comparable distribution of token inputs, is more appropriate. 
% Our model has only seen long documents during pretraining
This could be an explanation for why our base model is not able to beat the trained models in the short input length.\footnote{
% Longformer and BigBird have been warmstarted from RoBERTa (lots of short documents during this pretraining phase)
Note that Longformer and BigBird have been warm started from the RoBERTa checkpoint. Thus, they have been trained on short documents extensively during the first pretraining phase. Only in the second stage, these two models were fed long documents.
}

Despite the previous statement, we can also note that there is quite a clear correlation between the Micro-F1 and the number of parameters of the model in the case of small-size models. LegalBERT-small is an exception, outperforming DistilBERT but having fewer parameters. But LegalBERT-small has been pretrained on the same data as is contained in 6 out of 7 LexGLUE tasks. It is also likely, that the test sets have been contained in the pretraining data.
% We are still in this trend despite much less seen examples
Our small model is still in this trend of performance to model size, despite having seen much fewer examples during pretraining (almost 200 times fewer than BERT-Tiny). 
While in the case of the base model, this trend is still true for the same samples seen, if we leave out Legal-BERT and CaseLaw-BERT for the reasons already expressed.
This suggests that potentially extending the pretraining dataset with also short documents might improve the performance of our model in this regime as well. In our case, we avoided focusing too much on this point since the purpose of the paper is to solve the legal long documents as input.

% Another reason why we don't perform well is: we did not do any hyperparameter tuning at all
Finally, we did not tune the hyperparameters at all. 
It is well known that proper hyperparameter tuning and already selecting the right random seeds can significantly influence the downstream performance \cite{liu_empirical_2021, dodge_fine-tuning_2020}.
Note that especially our small models, like BERT-Tiny and miniLM, lag behind in the UnfairToS task (Macro-F1 score below 15). This could be due to an unlucky random seed (\citet{mosbach_stability_2021} and \citet{dodge_fine-tuning_2020} reported training performance strongly dependent on the random seed).

%\todo{Daniele, Joel: Find paper that showcases the effect of hyperparameter tuning: https://aclanthology.org/2021.acl-long.178.pdf, https://jmlr.org/papers/volume20/18-444/18-444.pdf}

%\todo{The models are very hard to compare: different num training steps, vocab, parameters. Can we somehow make them comparable?}

% Bert-Tiny: https://github.com/coastalcph/lex-glue/discussions/14
% miniLM: https://github.com/coastalcph/lex-glue/discussions/12
% DistilBERT: https://github.com/coastalcph/lex-glue/discussions/10
% LegalBERT: https://github.com/coastalcph/lex-glue/discussions/9

\section{Conclusions and Future Work}
\label{sec:conclusions_future_work}
%In this section, we answer the main research questions, discuss limitations, give a general conclusion and directions for future work.

\subsection{Answers to Main Research Questions}

\noindent \textbf{RQ1}: \emph{Is it possible to generate an ad-hoc \ac{LM} with domain (e.g., legal) expertise from scratch, reducing costs and CO$_2$ emissions?}
Yes, we showcase in this work that it is possible to pretrain a domain-expertise \ac{LM} from scratch with minimal compute, achieving comparable performance with methods that have seen more than three orders of magnitude more pretraining examples. Especially when there is no well-performing large teacher model available, our method is advisable.

\noindent \textbf{RQ2}: \emph{Is it possible to pretrain a Longformer model with the \ac{RTD} task (aka BudgetLongformer)?}
Yes, in this work, we show that it is possible to pretrain a Longformer model with the \ac{RTD} task.
% Can we write something more here?

\noindent \textbf{RQ3}: \emph{How does our BudgetLongformer compare with other models on the challenging summarization task? Particularly in the case of a legal domain-specific benchmark such as BillSum?}
Our \acp{LM} compare favorably to baselines on the challenging domain-specific summarization benchmark BillSum, requiring the models to process long inputs. Our small model outperforms the larger PEGASUS-base model, and our base model almost reaches the performance of the larger PEGASUS-large model. Both baselines have been pretrained with much more compute and data, and additionally with a pretraining task crafted specifically for summarization.

\noindent \textbf{RQ4}: \emph{ How well does our BudgetLongformer generalize to other domains, for example in the biomedical domain, as evaluated by the PubMed summarization benchmark?}
Yes, our results on the out-of-domain PubMed summarization benchmark show that our models compare favorably to baselines. Again, our small model outperforms PEGASUS-base and our base model approaches the performance of PEGASUS large.

\noindent \textbf{RQ5}: \emph{How do our \acp{LM} compare with other models on the understanding classification benchmark LexGLUE?}
Our small models compare favorably to baselines in their respective parameter range. Our base models approach the performance of the baselines even though (a) we trained using significantly less compute, (b) we did not pretrain on short documents, and (c) we did not tune the hyperparameters at all.

\subsection{Limitations}
\label{sec:limitations}

% ELECTRA cannot be warm started from already available checkpoints (e.g. longformer)
ELECTRA-style training has the disadvantage of the setup being slightly more complicated, requiring a generator and a discriminator. Additionally, the generator should be smaller than the discriminator to ensure stable training. This makes it difficult to warm start from available checkpoints, since two models of different sizes are required. Often, small models are not released, which makes it difficult to warm-start base models using the \ac{RTD} task. We leave the direction of warm starting a large discriminator with a base generator to future work.

% With the exception of EUR-LEX our model was only pretrained on US data
Except for EUR-LEX (1.31 GB or 1.8\% of our diverse dataset), our models have only seen US data during the pretraining phase. So, while these models are expected to work well on US data or datasets with similar content such as heavily influenced by the US or mainly common-law based, legal data from Europe for example is expected to look very different (mainly civil-law based except for the UK) and often translated from the original European languages. Thus, our models are not expected to transfer well to such kind of data.

% Because of lacking compute, we were not able to scale up our models and evaluate them on more downstream datasets
Because of insufficient compute, we were not able to scale up our models in terms of parameter size, batch size and number of pretraining steps. So while we can show that our approach scales well from the small to the base model, it is unknown if this continues to even larger model sizes. Although it is expected to produce better results, we do not know if using a higher batch size and more pretraining steps boosts performance significantly.
Additionally, the lacking compute budget made evaluating on more and especially large datasets like BigPatent impossible. Therefore, we cannot give any conclusions at this point to whether our results are robust across a wide range of datasets.

% We did not evaluate using other metrics such as BERTScore or BARTScore
So far, we did not evaluate our summarization models using newer metrics such as BERTScore \cite{zhang_bertscore_2020} or BARTScore \cite{yuan_bartscore_2021}. However, our baselines only evaluated using ROUGE, so we would have needed to rerun the baseline experiments to be able to compare our results to on these newer scores.

% No human evaluations on summarization datasets
So far, we did not have the resources to conduct a thorough human expert evaluation of the quality of our summarization outputs. Such an evaluation would be needed for production systems and for better comparison of models. However, it also requires highly educated medical experts (for PubMed) or lawyers with specific expertise in US bills (for BillSum) respectively, and thus a prohibitively high amount of resources.

% We compare based on seen examples, but not based on FLOPS.
For comparing the efficiency of pretraining, the number of FLOPs would probably be best. We compared the models' efficiency based on the number of seen examples during pretraining, due to ready availability (most papers report the batch size and the number of steps, but few papers report the FLOPs). \citet{liu_roberta_2019} for example, also report the number of GPU days used which we can also compare to. \citet{devlin_bert_2019}, however, trained using TPUs, which makes the comparison difficult again.

% We also did not evaluate reference-free evaluation metrics such as DAE or FastCC

\subsection{Conclusion}
% Main points:
% - very little pretraining works well
% - works very well for BillSum
% - given very little pretraining, we are very good already
% - LexGLUE is not our focus (because mostly short context), but we tested for completeness and we are ok, without tuning hyperparameters that are probably optimized for Legal-BERT

% longformer electra combo works
In this work, we show that we can successfully pretrain Longformer models with the \ac{RTD} task. 
% sota performance using very little pretraining on BillSum
Using very little pretraining we can achieve \ac{SOTA} performance on the challenging legal summarization task BillSum, outperforming PEGASUS, that has been pretrained specifically for summarization.
% also outperform Pegasus on out-of-domain PubMed dataset.
Our model even outperforms PEGASUS on the out-of-domain PubMed dataset involving biomedical research articles.
% We present a new way of getting a well performing language model with minimal compute and easy setup (no distillation required)
To sum up, we present a simple and extremely cheap way of pretraining a long-context \ac{LM} in cases without the availability of a large teacher model.

\subsection{Future Work}

% Investigate other datasets
Future work could test these models on further legal downstream tasks such as CUAD \cite{hendrycks_cuad_2021} or the recently released MultiLexSum \cite{shen_multi-lexsum_2022}. Additionally, one can test whether the out-of-domain results hold on other out-of-domain summarization datasets, such as BigPatent \cite{sharma_bigpatent_2019} or ArXiv \cite{cohan_discourse-aware_2018}.

% Scale up the models (batch size, num steps, num parameters and data)
Future work could further scale up the models in terms of batch size, number of pretraining steps, number of parameters and amount of data to test what further gains can be achieved.

% Tune hyperparams for LexGLUE
Due to compute constraints, we were unable to train the models long enough to reach \ac{SOTA} performance on LexGLUE. Future work could take our approach further and investigate the performance to be gained by investing more compute.

% investigate how to warm start ELECTRA from Longformer checkpoints (e.g. base model as generator and large model as discriminator)
Additionally, to save even more compute and to produce better models, one could investigate how to warm-start an ELECTRA pretraining from existing checkpoints. The difficulty, of course, lies in getting a suitable generator and discriminator trained with the same tokenizer. One possible setup might be Longformer-base as the generator and Longformer-large as the discriminator.

% Try other efficient transformers with electra pretraining
Finally, one can investigate the use of other efficient transformers with the \ac{RTD} task. 

%\section*{Acknowledgements}
% Write after the paper's acceptance. No citing of funding sources required.

\section*{Ethics Statement}
% There is no template for an Ethics Statement for TR. You could state that the work is on improving the training time of a large language model and there aren't any ethical implications by using the method (apart from reducing any carbon footprint significantly). Using LM in general, however, come with certain risk since the language produced could contain biases and could hurt protected groups. Researchers using the model should put into place respective safeguards to identify biased and/or toxic language.

% Carbon Footprint
Pretraining language models is a very compute-heavy process and thus leaves a large carbon footprint \cite{strubell_energy_2019, patterson_carbon_2021}. Our method makes significantly reduces the compute requirements and thus the carbon footprint. 

% Models can be biased and unfair
As with any large \ac{LM} there is the risk of it producing biased or unfair output. Researchers using the model should put into place respective safeguards to identify biased and/or toxic language.

% TODO motivate that we are crossing many discipline boundaries in AI: Papers on novel AI research problems, on AI techniques for novel application domains, and papers that cross discipline boundaries within AI are especially encouraged. (https://ijcai-23.org/call-for-papers/)

% TODO do error analysis based on franks comments:
%I would look at the summaries with a low ROUGE-score (and BERT-Score) and check for obvious discrepancies such as word duplications and/or hallucinations. It may also be instructive to compare examples where BudgetLongformer gets high scores but other approaches get lower scores and vice versa.
%If you select some summaries that look sensible to you but get an low automatic score, we may be able to find an in-house lawyer to get their opinion, but probably only for few examples.
%You could also correlated ROUGE and other automatic scores in order to get a better understanding of well the scores may work for these tasks.

\bibliography{bib/anthology,bib/references}
\bibliographystyle{styles/acl_natbib}

\appendix

\section{Overview of Compared Models}
\label{sec:models_overview}

\begin{table*}[t]
\centering
\resizebox{\textwidth}{!}{
    \begin{tabular}{llrrrrrrrr}
    \toprule
        Model Name & Source & P. Steps (K) & P. BS & \# P. Examples (M) & \# Enc. Params (M) & Max Seq Len & Vocab Size (K) & PubMed Rouge-L & BillSum Rouge-L \\ 
    \midrule
        DOC + SUM & \cite{kornilova_billsum_2019} & 1000 & 256 & 256 & 340 & 512 & 30 & ~ & 33.73 \\ 
        Transformer-base & \cite{zhang_pegasus_2020} & ~ & ~ & ~ & 110 & 1024 & 96 & 19.02 & 30.98 \\ 
        PEGASUS-base & \cite{zhang_pegasus_2020} & 500 & 256 & 128 & 110 & 1024 & 96 & 25.23 & 37.78 \\ 
        PEGASUS-large-C4 & \cite{zhang_pegasus_2020} & 500 & 8192 & 4096 & 340 & 1024 & 96 & 27.69 & 45.8 \\ 
%        BudgetLongformer small caselaw & ours & 100 & 32 & 3.2 & 29 & 4096 & 64 & ~ & 40.22 \\ 
        BudgetLongformer small diverse & ours & 100 & 32 & 3.2 & 29 & 4096 & 64 & 23.24 & 37.58 \\ 
        %BudgetLongformer small diverse 1024 -> 128 & ours & 100 & 32 & 3.2 & 29 & 4096 & 64 & 26.03 & 42.50 \\ 
%        BudgetLongformer base caselaw & ours & 100 & 32 & 3.2 & 159 & 4096 & 64 & ~ & 45.17 \\ 
        BudgetLongformer base diverse & ours & 100 & 32 & 3.2 & 159 & 4096 & 64 & 26.53 & 40.50 \\ 
        %BudgetLongformer base diverse 1024 -> 128 & ours & 100 & 32 & 3.2 & 159 & 4096 & 64 & 27.21 & 44.21 \\ 
    \end{tabular}
}
\caption{Abbreviations: P.: Pretraining, BS: Batch Size, Enc.: Encoder, Params: Parameters.
Comparison of the models evaluated on the summarization tasks BillSum and PubMed.
}
\label{tab:summarization_models}
\end{table*}

\begin{table*}[t]
\centering
\resizebox{\textwidth}{!}{
\begin{tabular}{llrrrrrrrrrrr}
    \toprule
        Model Name & Source & P. Steps (K) & P. BS & D. Steps (K) & D. BS & WS Steps (K) & WS BS & \# P. Examples (M) $\downarrow$ & \# Params (M) $\downarrow$ & Max Seq Len $\uparrow$ & Vocab Size (K) & LexGLUE Micro-F1 $\uparrow$ \\ 
    \midrule
    small models \\
    \midrule
        BERT-Tiny & \cite{turc_well-read_2019} & 1000 & 256 & 1400 & 256 & ~ & ~ & 614.4 & 4.4 & 512 & 31 & 70.1 \\ 
        miniLM & \cite{wang-etal-2021-minilmv2} & 1000 & 256 & 400 & 256 & ~ & ~ & 358.4 & 21 & 512 & 30 & 72.8 \\ 
        DistilBERT & \cite{sanh_distilbert_2020} & 1000 & 256 & 500 & 256 & ~ & ~ & 384 & 66 & 512 & 30 & 75.2 \\ 
        LegalBERT-small & \cite{chalkidis_legal-bert_2020} & 1000 & 256 & ~ & ~ & ~ & ~ & 256 & 35 & 512 & 31 & 76.7 \\ 
        BudgetLongformer small caselaw & ours & 100 & 32 & ~ & ~ & ~ & ~ & 3.2 & 29 & 4096 & 64 & 73.9 \\ 
        BudgetLongformer small diverse & ours & 100 & 32 & ~ & ~ & ~ & ~ & 3.2 & 29 & 4096 & 64 & 73.4 \\ 
        \midrule
        base models \\
        \midrule
        BERT & \cite{devlin_bert_2019} & 1000 & 256 & ~ & ~ & ~ & ~ & 256 & 110 & 512 & 30 & 77.8 \\ 
        RoBERTa & \cite{liu_roberta_2019} & 500 & 8192 & ~ & ~ & ~ & ~ & 4096 & 125 & 512 & 31 & 77.8 \\ 
        DeBERTa & \cite{he_debertav3_2021} & 1000 & 256 & ~ & ~ & ~ & ~ & 256 & 139 & 512 & 128 & 78.3 \\ 
        BigBird & \cite{zaheer_big_2021} & 500 & 8192 & ~ & ~ & 500 & 256 & 4224 & 127 & 4096 & 50 & 78.2 \\ 
        Longformer & \cite{beltagy_longformer_2020} & 500 & 8192 & ~ & ~ & 65 & 64 & 4100.16 & 149 & 4096 & 31 & 78.5 \\ 
        Legal-BERT-base & \cite{chalkidis_legal-bert_2020} & 1000 & 256 & ~ & ~ & ~ & ~ & 256 & 110 & 512 & 31 & 79.8 \\ 
        CaseLaw-BERT & \cite{zheng_when_2021} & 2000 & 256 & ~ & ~ & ~ & ~ & 512 & 110 & 512 & 30 & 79.4 \\ 
        BudgetLongformer base caselaw & ours & 100 & 32 & ~ & ~ & ~ & ~ & 3.2 & 159 & 4096 & 64 & 76.0 \\ 
        BudgetLongformer base diverse & ours & 100 & 32 & ~ & ~ & ~ & ~ & 3.2 & 159 & 4096 & 64 & 76.9 \\ 
    \bottomrule
\end{tabular}
}
\caption{Abbreviations: P.: Pretraining, D.: Distillation, WS: Warm Start, BS: Batch Size, Params: Parameters.
Comparison of the models evaluated on LexGLUE. In cases where we were not able to find the batch size in the papers, we assumed it to be 256, since this is the most widely used batch size in pretraining and the default for BERT. For DistilBERT we were not able to find the number of distillation steps, so we assumed 500K steps.
%DistilBERT was trained for 90 hours on 8 16GB NVIDIA V100 GPUs. RoBERTa was trained on 1024 32GB NVIDIA V100 GPUs for 500K steps. Our training took 60 hours on 4 16GB NVIDIA V100 GPUs. 
}
\label{tab:lexglue_models}
\end{table*}

In this section, we show detailed overviews of the model specifics (Tables \ref{tab:summarization_models} and \ref{tab:lexglue_models}).

\section{Detailed Results}
\label{sec:detailed_results}

\begin{table*}[t]
\centering
\resizebox{\textwidth}{!}{
    \begin{tabular}{lrrrr}
        \toprule
        Model (max-in-len->max-gen-len)                                          & \# Enc. Params $\downarrow$   & Rouge-1 $\uparrow$ & Rouge-2 $\uparrow$ & Rouge-L $\uparrow$ \\ 
        \midrule
%        Oracle \cite{kornilova_billsum_2019}                        & 45.11 & 28.74 & 37.38 \\ 
%        SumBasic \cite{kornilova_billsum_2019}                      & 30.74 & 14.16 & 23.92 \\ 
%        LSA \cite{kornilova_billsum_2019}                           & 32.64 & 15.69 & 26.26 \\ 
%        TextRank \cite{kornilova_billsum_2019}                      & 34.35 & 17.77 & 27.80 \\ 
%        DOC \cite{kornilova_billsum_2019}                           & 38.51 & 21.38 & 31.49 \\ 
%        SUM \cite{kornilova_billsum_2019}                           & 40.69 & 23.88 & 33.65 \\ 
        DOC + SUM (BERT large)                                & 340M      & 40.80 & 23.83 & 33.73 \\ 
        Transformer base                                     & 110M      & 44.05 & 21.30 & 30.98 \\ 
        PEGASUS base                                           & 110M      & 51.42 & 29.68 & 37.78 \\ 
        PEGASUS large (C4)                                    & 468M      & 57.20 & 39.56 & 45.80 \\ 
        PEGASUS large (HugeNews)                            & 468M      & 57.31 & 40.19 & 45.82 \\ 
%        BudgetLongformer small caselaw lr 5e-5                        & 29M & 48.93 & 26.90 & 36.98 \\  % 100K
%        BudgetLongformer small caselaw                                 & 29M & 51.62 & 30.84 & 40.22 \\  % 100K
%        BudgetLongformer small caselaw 200K                           & 29M & 49.02 & 27.02 & 36.98 \\ 
        BudgetLongformer small diverse (1024->128)                      & 29M & 53.61 & 33.54 & 42.50 \\  % 100K
        BudgetLongformer small diverse (1024->256)                       & 29M & 49.85 & 29.63 & 37.58 \\  % 100K
%        BudgetLongformer small diverse 200K                           & 29M & XXXX & XXXX & XXXX \\ 
%        BudgetLongformer base caselaw                                   & 159M & 56.10 & 36.50 & 45.17 \\  % 100K
%        BudgetLongformer base caselaw 200K                            & 159M & 55.30 & 35.47 & 44.30 \\ 
        BudgetLongformer base diverse (1024->256)                        & 159M & 52.70 & 32.97 & 40.50 \\  % 100K
        BudgetLongformer base diverse (1024->128)                        & 159M & 54.87 & 35.63 & 44.21 \\  % 100K
%        BudgetLongformer base diverse 200K                            & 159M & XXXX & XXXX & XXXX \\ 
        BudgetLongformer base diverse (4096->1024)                                  & 159M & 55.45 & 36.68 & 43.23 \\  % 100K
        \bottomrule
    \end{tabular}
}
\caption{Results on the BillSum dataset. Enc. Params is short for Encoder Parameters. }
\label{tab:billsum_results}
\end{table*}

\begin{table*}[t]
\centering
\resizebox{\textwidth}{!}{
    \begin{tabular}{lrrrr}
        \toprule
        Model  (max-in-len->max-gen-len)                                      & \# Enc. Params $\downarrow$   & Rouge-1 $\uparrow$ & Rouge-2 $\uparrow$ & Rouge-L $\uparrow$ \\ 
        \midrule
%        Previous SOTA TODO find citation                                & 340M      & 40.59 & 15.59 & 23.59  \\ 
        Transformer base                                        & 110M      & 33.94 & 7.43 & 19.02 \\ 
        PEGASUS base                                            & 110M      & 39.98 & 15.15 & 25.23 \\ 
        PEGASUS large (C4)                                     & 468M      & 45.49 & 19.90 & 27.69 \\ 
        PEGASUS large (HugeNews)                               & 468M      & 45.09 & 19.56 & 27.42 \\ 
%        BudgetLongformer small caselaw                                 & 29M & XXXX & XXXX & XXXX \\  % 100K
%        BudgetLongformer small caselaw 200K                           & 29M & XXXX & XXXX & XXXX \\ 
%        BudgetLongformer small diverse (1024->128)                         & 29M & 37.64 & 15.72 & 26.03 \\  % 100K
        BudgetLongformer small diverse (4096->512)                      & 29M & 34.98 & 13.56 & 23.24 \\  % 100K
%        BudgetLongformer small diverse 200K                           & 29M & XXXX & XXXX & XXXX \\ 
%        BudgetLongformer base caselaw                                   & 159M & XXXX & XXXX & XXXX \\  % 100K
%        BudgetLongformer base caselaw 200K                            & 159M & XXXX & XXXX & XXXX \\ 
%        BudgetLongformer base diverse (1024->128)                         & 159M & 39.19 & 16.93 & 27.21 \\ % 100K
        BudgetLongformer base diverse (4096->512)                        & 159M & 41.16 & 18.15 & 26.53 \\ % 100K
%        BudgetLongformer base diverse 200K                            & 159M & XXXX & XXXX & XXXX \\ 
        \bottomrule
    \end{tabular}
}
\caption{Results on the PubMed dataset. Enc. Params is short for Encoder Parameters.
% The caselaw models did not converge even when we tried different learning rates 1e-5, 3e-5 and 5e-5.
% The small caselaw model might converge with 3e-5. But we didn't finish the experiment due to cost constraints.
}
\label{tab:pubmed_results}
\end{table*}

\begin{table*}[t]
\centering
\resizebox{\textwidth}{!}{
    \begin{tabular}{lrrrrrrrrrrrr}
    \toprule
    model           & ECtHR A       & ECtHR B       & SCOTUS        & EUR-LEX       & LEDGAR        & UNFAIR-ToS    & CaseHOLD  & Average \\ 
    \midrule
    small models\\
    \midrule
    BERT-Tiny       & 63.7 / 44.0	& 63.9 / 50.4	& 61.1 / 35.7	& 57.9 / 25.0	& 83.8 / 73.3	& 93.9 / 11.1	& 66.2      & 70.1 / 43.7\\
    miniLM          & 67.9 / 55.1	& 66.6 / 61.0	& 60.8 / 45.5	& 62.2 / 35.6	& 86.7 / 79.6	& 93.9 / 13.2	& 71.3      & 72.8 / 51.6\\
    DistilBERT      & 69.9 / 61.1	& 70.5 / 69.1	& 67.0 / 55.9	& 66.0 / 51.5	& 87.5 / 81.5	& 97.1 / 79.4	& 68.6      & 75.2 / 66.7\\
    LegalBERT-small & 70.4 / 62.6*	& 71.3 / 69.4*	& 71.3 / 59.7*	& 66.1 / 48.2*	& 87.8 / 82.0*	& 97.4 / 81.7	& 72.9*     & 76.7 / 68.1\\
    %small caselaw 100K (3 seeds: 1,2,3)  & 64.8 / 46.7   & 75.2 / 58.6   & 70.5 / 50.0*  & 60.1 / 27.7   & 85.8 / 76.7   & 89.3 / 10.5   & 71.9* & 73.9 / 48.9\\
    BudgetLongformer small caselaw & 65.0 / 46.4 & 75.3 / 58.2 & 70.6 / 50.8*& 58.1 / 24.2 & 85.5 / 76.7 & 89.5 / 10.5 & 71.9* & 73.7 / 48.4\\  % 100K
%    small caselaw 200K (1 seed: 1)  & 61.6 / 38.2 & 76.5 / 51.2 & 71.6 / 50.2 & 56.9 / 22.4 & 84.6 / 74.3 & 89.5 / 10.5 & 72.3  & 73.3 / 45.6\\
    BudgetLongformer small diverse & 64.3 / 47.1 & 74.4 / 49.4 & 68.3 / 45.6* & 61.5 / 30.8* & 85.5 / 76.7* & 88.9 / 10.5 & 70.8*  & 73.4 / 47.3\\  % 100K
%    small diverse 200K (1 seed: 1)  & XXX / XXX   & XXX / XXX   & XXX / XXX  & XXX / XXX   & XXX / XXX  & XXX / XXX   & XXX     & XXX / XXX\\
    \midrule
    base models\\
    \midrule
    BERT                & 71.2 / 63.6 & 79.7 / 73.4 & 68.3 / 58.3 & 71.4 / 57.2 & 87.6 / 81.8 & 95.6 / 81.3 & 70.8 & 77.8 / 69.5\\
    RoBERTa             & 69.2 / 59.0 & 77.3 / 68.9 & 71.6 / 62.0 & 71.9 / 57.9 & 87.9 / 82.3 & 95.2 / 79.2 & 71.4 & 77.8 / 68.7\\
    DeBERTa             & 70.0 / 60.8 & 78.8 / 71.0 & 71.1 / 62.7 & 72.1 / 57.4 & 88.2 / 83.1 & 95.5 / 80.3 & 72.6 & 78.3 / 69.7\\
    BigBird             & 70.0 / 62.9 & 78.8 / 70.9 & 72.8 / 62.0 & 71.5 / 56.8 & 87.8 / 82.6 & 95.7 / 81.3 & 70.8 & 78.2 / 69.6\\
    Longformer          & 69.9 / 64.7 & 79.4 / 71.7 & 72.9 / 64.0 & 71.6 / 57.7 & 88.2 / 83.0 & 95.5 / 80.9 & 71.9 & 78.5 / 70.5\\
    CaseLawBERT         & 69.8 / 62.9 & 78.8 / 70.3 & 76.6 / 65.9* & 70.7 / 56.6 & 88.3 / 83.0 & 96.0 / 82.3 & 75.4* & 79.4 / 70.9\\
    LegalBERT-base      & 70.0 / 64.0* & 80.4 / 74.7* & 76.4 / 66.5* & 72.1 / 57.4* & 88.2 / 83.0* & 96.0 / 83.0 & 75.3* & 79.8 / 72.0\\
    BudgetLongformer base caselaw  & 67.2 / 55.9 & 76.6 / 61.1 & 74.9 / 62.3* & 64.7 / 42.9 & 86.9 / 80.4 & 89.5 / 10.5 & 72.1*  & 76.0 / 55.0\\ % 100K
%    base caselaw 200K  (1 seed: 1)  & 69.0 / 55.4 & 77.7 / 62.8 & 73.1 / 61.7 & 65.0 / 42.7 & 87.3 / 81.6  & 88.9 / 10.5 & 71.9 & 76.1 / 55.2\\
    BudgetLongformer base diverse & 66.3 / 52.6 & 77.9 / 72.3 & 75.4 / 62.9* & 65.6 / 44.4* & 87.0 / 81.0*  & 95.1 / 76.7 & 71.3* & 76.9 / 65.9\\  % 100K 
%    base diverse 200K  (1 seed: 1)  & XXX / XXX   & XXX / XXX   & XXX / XXX  & XXX / XXX   & XXX / XXX  & XXX / XXX   & XXX     & XXX / XXX\\
    \bottomrule
    \end{tabular}
}
\caption{Results on LexGLUE. Because of limited compute, we only ran 1 random seed for our models. The other results are reported on GitHub\footnote{\url{https://github.com/coastalcph/lex-glue/discussions/categories/new-results}}. The asterix denotes datasets which are (partly) covered in the pretraining dataset. For each column we report the results in the format micro-averaged F1 score / macro-average F1 score. For the CaseHOLD task, both scores are the same.
%\todo{Mention that we did not tune the hyperparameters and the hyperparams are probably tuned for Legal-BERT}
}
\label{tab:lexglue_results}
\end{table*}

\begin{figure}[ht]
    \centering
    \resizebox{\columnwidth}{!}{
        \includegraphics{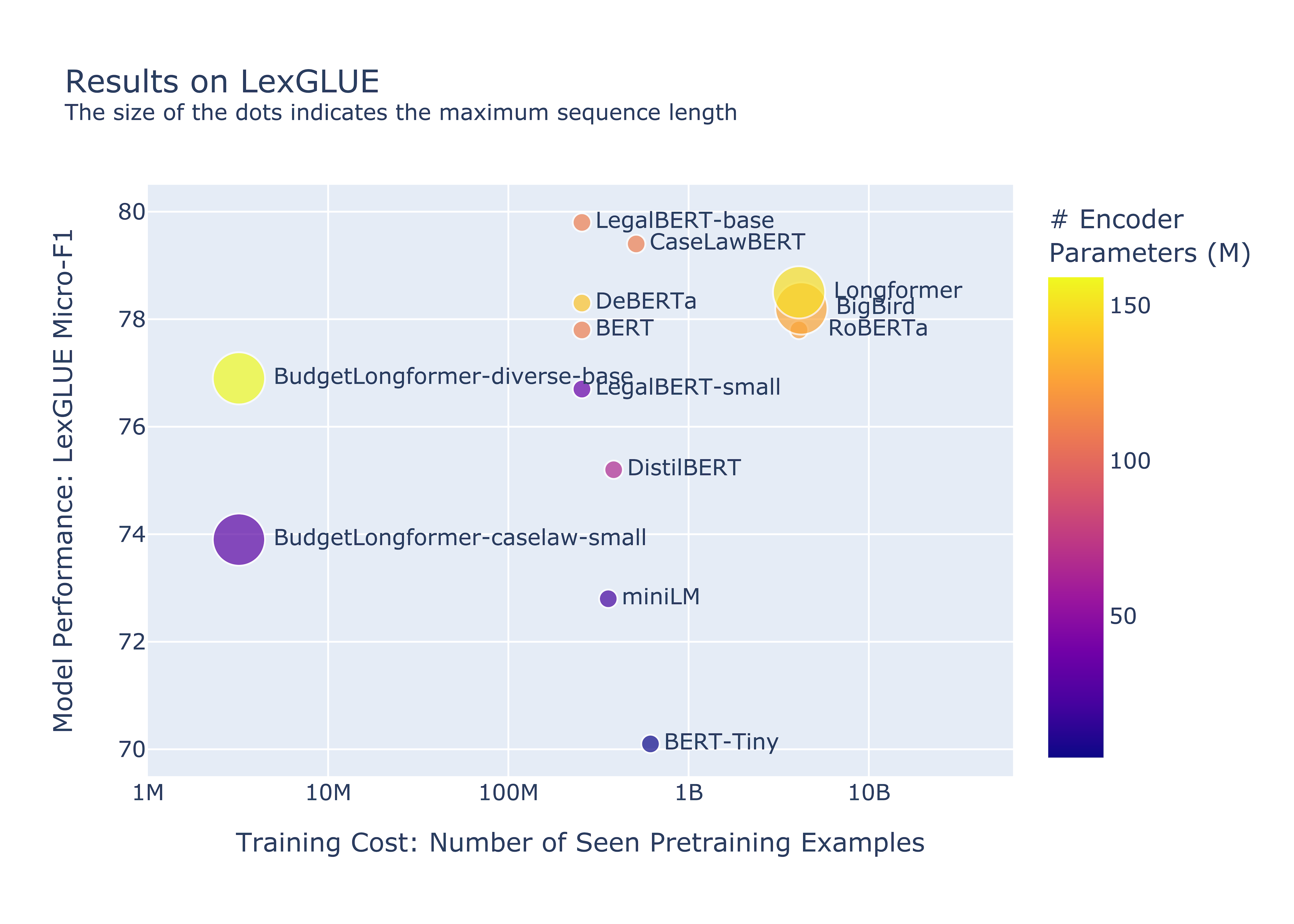}
    }
    \vspace{-7mm}
    \caption{Results on the LexGLUE benchmark (all models). Note that the x-axis is in log-scale.}
    \label{fig:lexglue}
    \vspace{-5mm}
\end{figure}

In this section, we show detailed and comprehensive results of the compared models (Tables \ref{tab:billsum_results}, \ref{tab:pubmed_results} and \ref{tab:lexglue_results} and Figure \ref{fig:lexglue}).

\section{Pretraining Details}
\label{sec:pretraining_details}

\begin{table}[t]
\centering
\resizebox{\columnwidth}{!}{
    \begin{tabular}{llrrr}
    \toprule
    Model & Data    & \# Steps  & Train Loss  & Eval Loss \\ 
    \midrule
    small & caselaw &  50K  &   14.61 &     15.78 \\
    small & caselaw & 100K  &   13.93 &     15.07 \\
%    small & caselaw & 150K  &   13.63 &     14.77 \\
%    small & caselaw & 200K  &   13.38 &     XX.XX \\
    \midrule
    small & diverse &  50K  &   13.75 &     12.70 \\
    small & diverse & 100K  &   12.78 &     11.66 \\
%    small & diverse & 150K  &   12.28 &     11.29 \\
%    small & diverse & 200K  &   12.05 &     11.03 \\
    \midrule
    base  & caselaw &  50K  &   12.40 &     13.76 \\
    base  & caselaw & 100K  &   11.67 &     12.99 \\
%    base  & caselaw & 150K  &   11.31 &     12.58 \\
%    base  & caselaw & 200K  &   11.02 &     12.27 \\
    \midrule
    base  & diverse &  50K  &   10.70 &     10.01 \\
    base  & diverse & 100K  &    9.86 &      9.22 \\
%    base  & diverse & 150K  &   9.42 &     8.79 \\
%    base  & diverse & 200K  &   9.20 &     8.56 \\
    \bottomrule
    \end{tabular}
}
\caption{Training and Evaluation losses for the different trained models. Note that these losses are the addition of the loss of the generator and the loss of the discriminator. Since the loss of the discriminator is much smaller, it is scaled by a factor of 50 to stabilize training.}
\label{tab:losses}
\end{table}

In this section, we show additional details regarding the pretraining process (Table \ref{tab:losses}).

\section{Hyperparameters and Training Details}
\label{sec:hyperparameters}

In this section, we present additional details regarding the chosen hyperparameters.

\subsection{Pretraining}

We pretrained our models with batch size 32 and learning rate 5e-4 and 3e-4 for the small and base models respectively. 
We used a Longformer attention window of 256. % To fit it into the GPU memory
As described in by \citet{clark_electra_2020}, we used 10000 warm up steps and a 4 and 3 times smaller generator than the discriminator in the small and base version respectively. In contrast to \citet{clark_electra_2020} we reduced the generator's depth (number of hidden layers) instead of its width (embedding size, hidden size and intermediate size). 
We used a \ac{MLM} probability of 25\% for the generators.

For running the pretraining, we used an AWS p3.8xlarge instance with 4 16GB NVIDIA V100 GPUs. Training the four models to 100K steps each, took approx. 18 days or 72 GPU days in total. Previous debug runs additionally consumed approx. 3 days or 12 GPU days.

\subsection{Downstream Benchmarks}

% diverse models were more robust 
Overall, we found the diverse models to be more robust in finetuning with less failed runs and typically higher performance.

For running the finetuning experiments, we used an AWS p3.16xlarge instance with 8 16GB NVIDIA V100 GPUs. Running the BillSum, PubMed, and LexGLUE experiments including hyperparameter tuning took approximately 25, 7, and 11 GPU days in total respectively.

\section{Library Versions}
We used the following versions to the libraries in a pip requirements.txt format:\\
datasets==2.4.0\\
huggingface-hub==0.9.0\\
nltk==3.7\\
pandas==1.3.5\\
rouge-score==0.1.2\\
scikit-learn==1.0.2\\
scipy==1.7.3\\
tokenizers==0.12.1\\
torch==1.12.1\\
tqdm==4.64.0\\
transformers==4.21.1\\

\section{Examples}
Example summaries are displayed in Tables \ref{tab:examples_billsum-1024-256_small_diverse}, \ref{tab:examples_billsum-1024-256_base_diverse}, \ref{tab:examples_billsum-4096-1024_base_diverse}, \ref{tab:examples_pubmed-4096-512_small_diverse}, and \ref{tab:examples_pubmed-4096-512_base_diverse}. Since the documents are very long sometimes, we truncated them to the first 2500 characters. We sorted the examples by RougeL scores and show the bottom 5\%, bottom 25\%, top 75\% and top 95\% percentile. 
% We notice, that the summaries are often stopped early, because the maximum generation length is set to 128 tokens. 

%\todo{Maybe it would make more sense to sort by Rouge1 score instead of RougeL}

\renewcommand{\arraystretch}{0.2} % reduces vertical whitespace in cells

\begin{table*}[t]
\centering
\resizebox{\textwidth}{!}{
    \begin{tabular}{ l p{21cm} }
    \toprule
\\ 
 \midrule 
\multicolumn{2}{l}{Bottom 5\% example (Sorted by rougeL)}\\ 
 \midrule 
Document & \tiny .--(1) For purposes 
of subsection (a)(2) and this subsection, the term `joint resolution' 
means only a joint resolution introduced by a qualifying Member 
specified in paragraph (2) after the date on which the report of the 
President under subsection (a)(1) is received by the Congress--
            ``(A) the matter after the resolving clause of which is as 
        follows: `That the Congress hereby concurs in the certification 
        of the President relating to deployment of a National Missile 
        Defense system as submitted to Congress pursuant to section 
        4(b) of the National Missile Defense Act of 1999.';
            ``(B) which does not have a preamble; and
            ``(C) the title of which is as follows: `Joint resolution 
        relating to deployment of a National Missile Defense system.'.
    ``(2) For purposes of this subsection, a qualifying Member 
described in this paragraph is--
            ``(A) in the case of the House of Representatives, the 
        majority leader or minority leader of the House of 
        Representatives or a Member of the House of Representatives 
        designated by the majority leader or minority leader; and
            ``(B) in the case of the Senate, the majority leader or 
        minority leader of the Senate or a Member of the Senate 
        designated by the majority leader or minority leader.
    ``(3) The provisions of paragraphs (3) through (8) of section 4(c) 
of the National Missile Defense Deployment Criteria Act of 2001 shall 
apply to a joint resolution under this subsection in the same manner as 
to a joint resolution under such section.''.

SEC. 4. LIMITATION ON OBLIGATION OF FUNDS FOR PROCUREMENT FOR NATIONAL 
              MISSILE DEFENSE SYSTEM.

    (a) Limitation.--No funds appropriated to the Department of Defense 
for procurement may be obligated for the National Missile Defense 
system unless--
            (1) the President submits to Congress a report concerning 
        testing of the National Missile Defense system against 
        countermeasures that includes a certification described in 
        subsection (b); and
            (2) a joint resolution concurring in the President's 
        certification in such report is enacted as provided for in this 
        section.
    (b) Presidential Certification.--A certification described in this 
subsection is a certification by the President that--
            (1) an adequate testing program for the National Missile 
... \\ 
 \midrule 
Gold & \tiny National Missile Defense Deployment Criteria Act of 2001 - Amends the National Missile Defense Act of 1999 to allow deployment of a national missile defense system (system) only if: (1) the system is technologically feasible; (2) system cost in relation to other Department of Defense (DOD) priorities will not lead to an overall reduction in national security by reducing resources available for other defense priorities; (3) the system will not diminish overall U.S. national security; (4) the system will not threaten to disrupt relations with U.S. nuclear allies, U.S. European allies, Russia, the People's Republic of China, and other nations; and (5) the threat of a long-range ballistic missile attack from a nation of concern is clearly demonstrated.Prohibits the President from directing DOD to deploy a system unless and until: (1) the President certifies to Congress that the above deployment conditions have been met; and (2) a joint resolution is enacted concurring in the President's certification.Prohibits DOD procurement funds from being obligated for a system unless: (1) the President certifies to Congress that adequate system tests have been undertaken to meet identified threats against countermeasures; and (2) a joint resolution is enacted concurring in the President's certification.Requires the Secretary of Defense to direct the Ballistic Missile Defense Organization to: (1) include specified system countermeasures in system ground and flight testing conducted before the system becomes operational; and (2) determine the extent to which the exoatmospheric kill vehicle and the system can reliably discriminate between warheads and such countermeasures. \\ 
 \midrule 
Model & \small prohibits funds appropriated to the department of defense (dod) for procurement from being obligated for the national missile defense system unless the president certifies to congress that: (1) an adequate testing program for the system is in place to meet the threats identified in the report; and (2) an adequate ground and flight testing of the system has been conducted against the system that are likely to be used against the system and that other countries have or are likely to acquire. \\ 
 \midrule 
Metrics & Rouge1: 40.69, Rouge2: 16.67, RougeL: 20.0, RougeLsum: 20.0, Summary length (tokens): 94
 \\ 
\\ 
 \midrule 
\multicolumn{2}{l}{Bottom 25\% example (Sorted by rougeL)}\\ 
 \midrule 
Document & \tiny TITLE I--FEDERAL AIRPORTS SECURITY ENHANCEMENT ACT

SEC. 101. SHORT TITLE.

    This title may be cited as the ``Federal Airports Security 
Enhancement Act''.

SEC. 102. ESTABLISHMENT OF AIRPORT SECURITY COMMITTEES.

    The Act of July 5, 1994 (49 U.S.C. 44935), is amended--
            (1) by striking section 44901 subparagraph (b) and 
        inserting the following:

``SEC. 103. EMPLOYMENT STANDARDS AND TRAINING.''.

            (2) by striking section 44935 subparagraph (b) and 
        inserting the following:
    ``(a) Review and Recommendations.--The Administrator of the Federal 
Aviation Administration shall establish Security Committees at each 
airport location to be composed of representatives of the air carriers, 
airport operators, other interested parties and at least one 
representative from the Federal Protective Service, the Federal Bureau 
of Investigation, The Federal Aviation Administration and one member 
from each local jurisdiction that the airport may be located in or that 
may have jurisdictional authority for the airport facility. Each 
Airport Security Committee shall meet at least quarterly and shall make 
recommendations for minimum security countermeasures to the 
Administrator. The Federal Protective Service shall have primary 
responsibility for conducting on an ongoing basis security surveys and 
formulating recommendations to the Security Committee. The 
Administrator shall prescribe appropriate changes in existing 
procedures to improve that performance.''.

SEC. 103. SCREENING PASSENGERS AND PROPERTY.

    The Act of July 5, 1994 (49 U.S.C. 44935), is amended by striking 
section 44901, subparagraph (a), and inserting the following:
    ``(a) General Requirements.--The Administrator of the Federal 
Aviation Administration shall prescribe regulations requiring screening 
of all passengers and property that will be carried in a cabin of an 
aircraft in air transportation or intrastate air transportation. The 
screening must take place before boarding and be carried out by a 
weapon detecting facility or procedure used or operated by an employee 
or agent of the Federal Protective Service. The Administrator--
            ``(1) shall require that sufficient Federal Police Officers 
        are posted at airport facilities to provide patrol duties 
        during all hours of operations as well as supervise screening 
        personnel;
            ``(2) shall maintain sufficient numbers of Special Agents 
        to provid... \\ 
 \midrule 
Gold & \tiny Federal Airports Security Enhancement Act - Amends Federal aviation law to direct the Administrator of the Federal Aviation Administration (FAA) to establish at each airport a Security Committee which shall make recommendations for minimum security counter-measures. Requires the Administrator, on the basis of such recommendations, to prescribe appropriate changes to improve the performance of existing airport security procedures.Requires the screening of passengers and property that will be carried in a cabin of an aircraft to be carried out by Federal Protective Service employees or agents. (Currently, screening is carried out by employees or agents of an air carrier, interstate air carrier, or foreign air carrier).Authorizes the Administrator of the General Services Administration (GSA) to appoint police officers and special agents (currently, special policemen and nonuniformed special policemen) for the policing of all Federal buildings (including buildings under the control of the GSA). Sets forth certain additional powers of such officers and agents, including the authority to carry firearms and to police areas adjacent to Federal property.Establishes the Federal Protective Service as a separate operating service of the GSA. Calls for at least 1,000 full-time equivalent Service police officers to be assigned to areas outside of airport operations. Requires the Commissioner of the Service to prescribe minimum employment and training standards to be applied in the contracting of security personnel for the policing of buildings and areas controlled by the United States and GSA. Authorizes GSA to recover airport security costs from the FAA. \\ 
 \midrule 
Model & \small table of contents: 
     title i:  federal airports security enhancement act 
     title ii:  miscellaneous provisions 
general federal airports security enhancement act - 
title i: federal airports security enhancement
 - amends the federal aviation act of 1992 to direct the administrator of the federal aviation administration (faa) to prescribe regulations requiring screening of all passengers and property that will be carried in a port of aircraft in air transportation or intrastate air transportation. 
(sec. 102) directs the administrator to prescribe regulations requiring screening of all passengers and property that will be carried out by the federal protective service, the federal bureau of investigation (fbi), the federal bureau of investigation (fbi), and one member from each local jurisdiction that the aircraft may be located in or that may have jurisdictional authority for the airport of an aircraft in air transportation or intrastate air transportation. 
(sec. 103) directs the administrator to prescribe regulations requiring screening of all passengers and property that will be carried out by a weapon detection facility or procedure used or operated by an employee or agent of the federal protective service. 
(sec. 103) authorizes the administrator to enter into agreements with state and local law enforcement authorities to obtain authority for, jointly with state and local law enforcement authorities. 
( \\ 
 \midrule 
Metrics & Rouge1: 52.44, Rouge2: 22.84, RougeL: 29.7, RougeLsum: 47.8, Summary length (tokens): 256
 \\ 
\\ 
 \midrule 
\multicolumn{2}{l}{Top 75\% example (Sorted by rougeL)}\\ 
 \midrule 
Document & \tiny SECTION 1. SHORT TITLE.

    This Act may be cited as the ``Patent and Trademark Office 
Authorization Act of 2002''.

SEC. 2. AUTHORIZATION OF AMOUNTS AVAILABLE TO THE PATENT AND TRADEMARK 
              OFFICE.

    (a) In General.--There are authorized to be appropriated to the 
United States Patent and Trademark Office for salaries and necessary 
expenses for each of the fiscal years 2003 through 2008 an amount equal 
to the fees estimated by the Secretary of Commerce to be collected in 
each such fiscal year, respectively, under--
            (1) title 35, United States Code; and
            (2) the Act entitled ``An Act to provide for the 
        registration and protection of trademarks used in commerce, to 
        carry out the provisions of certain international conventions, 
        and for other purposes'', approved July 5, 1946 (15 U.S.C. 1051 
        et seq.) (commonly referred to as the Trademark Act of 1946).
    (b) Estimates.--Not later than February 15, of each fiscal year, 
the Undersecretary of Commerce for Intellectual Property and the 
Director of the Patent and Trademark Office (in this Act referred to as 
the Director) shall submit an estimate of all fees referred to under 
subsection (a) to be collected in the next fiscal year to the chairman 
and ranking member of--
            (1) the Committees on Appropriations and Judiciary of the 
        Senate; and
            (2) the Committees on Appropriations and Judiciary of the 
        House of Representatives.

SEC. 3. ELECTRONIC FILING AND PROCESSING OF PATENT AND TRADEMARK 
              APPLICATIONS.

    (a) Electronic Filing and Processing.--Not later than December 1, 
2004, the Director shall complete the development of an electronic 
system for the filing and processing of patent and trademark 
applications, that--
            (1) is user friendly; and
            (2) includes the necessary infrastructure to--
                    (A) allow examiners and applicants to send all 
                communications electronically; and
                    (B) allow the Office to process, maintain, and 
                search electronically the contents and history of each 
                application.
    (b) Authorization of Appropriations.--Of amounts authorized under 
section 2, there are authorized to be appropriated to carry out 
subsection (a) of this section not more than \$50,000,000 for each of 
fiscal years 2003 and 2004. Amounts made available under this 
subsection shall... \\ 
 \midrule 
Gold & \tiny Patent and Trademark Office Authorization Act of 2002 - Authorizes appropriations to the U.S. Patent and Trademark Office for salaries and expenses for FY 2003 through 2008 in an amount equal to all patent and trademark fees estimated by the Secretary of Commerce (Secretary) to be collected in each such fiscal year.(Sec. 2) Requires the Under Secretary of Commerce for Intellectual Property and the Director of the Office (Director), by February 15 of each fiscal year, to report an estimate of all fees to be collected in the next fiscal year to the chairman and ranking member of specified congressional committees.(Sec. 3) Requires the Director, by December 1, 2004, to complete the development of an electronic system for the filing and processing of patent and trademark applications that: (1) is user friendly; and (2) includes the necessary infrastructure to allow examiners and applicants to send all communications electronically, and the Office to process, maintain, and search electronically the contents and history of each application. Authorizes appropriations for FY 2003 and 2004 for development of such system.(Sec. 4) Requires the Secretary, in each of the five calendar years following the enactment of this Act, to report to specified congressional committees on the progress made in implementing the 21st Century Strategic Plan issued on June 3, 2002, and on any amendments made to it.(Sec. 5) Amends Federal patent law to provide that previous citation by or to, or consideration by the Office of, a patent or printed publication does not preclude the existence of a substantial new question of patentability in patent reexamination proceedings.(Sec. 6) Revises requirements for appeals in inter partes reexamination proceedings to allow a third-party requester to appeal to the U.S. Court of Appeals for the Federal Circuit, or be a party to any appeal taken by the patent owner, with respect to any final decision favorable to the patentability of any original or proposed amended or new claim of the patent.Allows a third-party requester to appeal a decision of the Board of Patent Appeals and Interferences.Provides that a third-party requester in an inter partes reexamination proceeding dissatisfied with the final decision in an appeal to the Board may appeal the decision only to the U.S. Court of Appeals for the Federal Circuit. \\ 
 \midrule 
Model & \small patent and trademark office authorization act of 2002 - authorizes appropriations to the u.s. patent and trademark office for fy 2003 through 2008. requires the director of the patent and trademark office to: (1) complete the development of an electronic system for the filing and processing of patent and trademark applications; and (2) submit an annual report to the congressional committees on progress made in implementing the 21st century strategic plan issued under the federal patent and trademark programs. \\ 
 \midrule 
Metrics & Rouge1: 48.99, Rouge2: 39.86, RougeL: 44.3, RougeLsum: 48.32, Summary length (tokens): 94
 \\ 
\\ 
 \midrule 
\multicolumn{2}{l}{Top 95\% example (Sorted by rougeL)}\\ 
 \midrule 
Document & \tiny SECTION 1. SHORT TITLE.

    This Act may be cited as the ``Guidance, Understanding, and 
Information for Dual Eligibles (GUIDE) Act''.

SEC. 2. FINDINGS; PURPOSE.

    (a) Findings.--The Congress finds the following:
            (1) Nearly 8,800,000 Americans were eligible for benefits 
        under the Medicare program and for medical assistance under 
        Medicaid (dual eligible beneficiaries) in fiscal year 2005. Of 
        these ``dual eligible beneficiaries'', almost 40 percent have 
        cognitive impairments, including Alzheimer's disease, dementia, 
        serious mental illnesses, and intellectual disabilities. Until 
        December 31, 2005, dual eligible beneficiaries received 
        outpatient prescription drug benefits through medical 
        assistance under Medicaid. On January 1, 2006, drug coverage 
        for dual eligibles switched from Medicaid to Medicare.
            (2) In 2008, 53 percent of dual eligible beneficiaries had 
        medication access problems and of those, 27 percent experienced 
        significant adverse clinical events.
            (3) Individuals with medication access issues experience 
        significantly more adverse clinical events. Among dual eligible 
        beneficiaries with mental illness who had medication access 
        problems, 27 percent experienced significant adverse clinical 
        events, which included emergency room visits and 
        hospitalizations.
            (4) In total, over 1,000,000 dual eligible beneficiaries 
        and low-income subsidy beneficiaries were automatically auto-
        enrolled to new benchmark prescription drug plans under part D 
        of the Medicare program between 2006 and 2007.
            (5) Community providers are at the front line of helping 
        the most vulnerable dual eligible beneficiaries obtain 
        prescription drug coverage under the Medicare program and 
        navigate complex enrollment and low-income subsidy eligibility 
        requirements under such program.
    (b) Purpose.--It is the purpose of this bill to help low-income 
persons with cognitive impairments to enroll in and navigate the 
prescription drug benefit under the Medicare program by providing front 
line community providers who serve the population daily with financial 
assistance to conduct vigorous education and outreach and direct case 
management.

SEC. 3. MEDICARE PRESCRIPTION DRUG OUTREACH DEMONSTRATION PROGRAM FOR 
              DUAL E... \\ 
 \midrule 
Gold & \tiny Guidance, Understanding, and Information for Dual Eligibles (GUIDE) Act - Directs the Secretary of Health and Human Services to establish a three-year demonstration program under which the Secretary awards grants and contracts to appropriate, qualified community programs and clinics for individuals with intellectual or developmental disabilities, or certain programs under the Public Health Services Act, to employ qualified social workers and case managers to provide one-on-one counseling about benefits under part D (Voluntary Prescription Drug Benefit Program) of title XVIII (Medicare) of the Social Security Act (SSA) to a full-benefit dual eligible individual (eligible for benefits under both Medicare and SSA title XIX [Medicaid]) who has one or more mental disabilities. \\ 
 \midrule 
Model & \small guidance, understanding, and information for dual eligible beneficiaries with intellectual or developmental disabilities act - directs the secretary of health and human services (hhs) to establish a three-year demonstration program under which the secretary awards grants and contracts to qualified community programs and clinics for individuals with intellectual or developmental disabilities or such programs to provide medicare prescription drug assistance to individuals with intellectual or developmental disabilities or such programs. \\ 
 \midrule 
Metrics & Rouge1: 60.87, Rouge2: 47.25, RougeL: 58.7, RougeLsum: 58.7, Summary length (tokens): 80
 \\ 

    \bottomrule
    \end{tabular}
}
\caption{Examples of the BillSum dataset using the model billsum-1024-256 small diverse}
\label{tab:examples_billsum-1024-256_small_diverse}
\end{table*}

\begin{table*}[t]
\centering
\resizebox{\textwidth}{!}{
    \begin{tabular}{ l p{21cm} }
    \toprule
\\ 
 \midrule 
\multicolumn{2}{l}{Bottom 5\% example (Sorted by rougeL)}\\ 
 \midrule 
Document & \tiny SECTION 1. SHORT TITLE.

    This Act may be cited as the ``Health Coverage Tax Credit Extension 
Act of 2015''.

SEC. 2. EXTENSION AND MODIFICATION OF HEALTH COVERAGE TAX CREDIT.

    (a) Extension.--Subparagraph (B) of section 35(b)(1) of the 
Internal Revenue Code of 1986 is amended by striking ``before January 
1, 2014'' and inserting ``before January 1, 2020''.
    (b) Coordination With Credit for Coverage Under a Qualified Health 
Plan.--Subsection (g) of section 35 of the Internal Revenue Code of 
1986 is amended--
            (1) by redesignating paragraph (11) as paragraph (13), and
            (2) by inserting after paragraph (10) the following new 
        paragraphs:
            ``(11) Election.--
                    ``(A) In general.--A taxpayer may elect to have 
                this section apply for any eligible coverage month.
                    ``(B) Timing and applicability of election.--Except 
                as the Secretary may provide--
                            ``(i) an election to have this section 
                        apply for any eligible coverage month in a 
                        taxable year shall be made not later than the 
                        due date (including extensions) for the return 
                        of tax for the taxable year, and
                            ``(ii) any election for this section to 
                        apply for an eligible coverage month shall 
                        apply for all subsequent eligible coverage 
                        months in the taxable year and, once made, 
                        shall be irrevocable with respect to such 
                        months.
            ``(12) Coordination with premium tax credit.--
                    ``(A) In general.--An eligible coverage month to 
                which the election under paragraph (11) applies shall 
                not be treated as a coverage month (as defined in 
                section 36B(c)(2)) for purposes of section 36B with 
                respect to the taxpayer.
                    ``(B) Coordination with advance payments of premium 
                tax credit.--In the case of a taxpayer who makes the 
                election under paragraph (11) with respect to any 
                eligible coverage month in a taxable year or on behalf 
                of whom any advance payment is made under section 7527 
                with respect to any month in such taxable year--
                    ... \\ 
 \midrule 
Gold & \tiny Health Coverage Tax Credit Extension Act of 2015 This bill extends the tax credit for health insurance costs of a taxpayer and qualifying family members through 2019. The tax credit for health insurance costs is a refundable tax credit equal to 72.5\% of the cost of qualified health coverage paid by an eligible individual [defined as an individual who is receiving a trade adjustment allowance, is eligible for the alternative trade adjustment assistance program, or is over age 55 and receives pension benefits from the Pension Benefit Guaranty Corporation (PBGC)]. The bill requires a taxpayer to make an election to have the tax credit apply for any eligible coverage month during a taxable year. An eligible coverage month is a month in which an eligible individual is covered by qualified health insurance, does not have other specified coverage, and is not imprisoned.  The bill also directs the Departments of the Treasury, Health and Human Services, and Labor and the PBGC to conduct a public outreach, including on the Internet, to inform individuals eligible for the tax credit for health insurance costs on the extension of such credit and the availability of the election to claim such credit retroactively for coverage months beginning after December 31, 2013.  \\ 
 \midrule 
Model & \small health coverage tax credit extension act of 2015 this bill amends the internal revenue code, with respect to health care coverage, to: (1) extend through 2020 the tax credit for advance payments to individuals, (2) allow advance payments of advance payments of advance payments of advance payments, and (3) extend through 2018 the tax credit for advance payments of advance payments of advance payments to individuals. \\ 
 \midrule 
Metrics & Rouge1: 26.37, Rouge2: 11.07, RougeL: 21.25, RougeLsum: 25.64, Summary length (tokens): 82
 \\ 
\\ 
 \midrule 
\multicolumn{2}{l}{Bottom 25\% example (Sorted by rougeL)}\\ 
 \midrule 
Document & \tiny SECTION 1. EXTENSION.
    (a) In General.--Chapter 5 of subtitle B of the Agricultural 
Marketing Act of 1946 (7 U.S.C. 1636 et seq.) is amended by adding at 
the end the following new section:
    ``SEC. 260. TERMINATION OF AUTHORITY.
    ``The authority provided by this subtitle terminates on September 
30, 2010.''.
    (b) Conforming Amendment and Extension.--Section 942 of the 
Livestock Mandatory Reporting Act of 1999 (7 U.S.C. 1635 note; Public 
Law 106-78) is amended by striking ``terminate on September 30, 2005'' 
and inserting ``(other than section 911 of subtitle A and the 
amendments made by that section) terminate on September 30, 2010''.
SEC. 2. DEFINITIONS.
    (a) Base Market Hogs.--Section 231(4) of the Agricultural Marketing 
Act of 1946 (7 U.S.C. 1635i(4)) is amended to read as follows:
        ``(4) Base market hog.--The term `base market hog' means a 
    barrow or gilt for which no discounts are subtracted from and no 
    premiums are added to the base price.''.
    (b) Boars.--Section 231(5) of such Act (7 U.S.C. 1635i(5)) is 
amended to read as follows:
        ``(5) Boar.--The term `boar' means a sexually-intact male 
    swine.''.
    (c) Packer of Sows and Boars.--Section 231(12) of such Act (7 
U.S.C. 1635i(12)) is amended by--
        (1) striking subparagraph (B) and inserting the following new 
    subparagraph:
            ``(B) for any calendar year, the term includes only--
                ``(i) a swine processing plant that slaughtered an 
            average of at least 100,000 swine per year during the 
            immediately preceding five calendar years; and
                ``(ii) a person that slaughtered an average of at least 
            200,000 sows, boars, or any combination thereof, per year 
            during the immediately preceding five calendar years; 
            and''; and
        (2) in subparagraph (C)--
            (A) by inserting ``or person'' after ``swine processing 
        plant'';
            (B) by inserting ``or person'' after ``plant capacity of 
        the processing plant''; and
            (C) by inserting ``or person'' after ``determining whether 
        the processing plant''.
SEC. 3. REPORTING; BARROWS AND GILTS.
    Section 232(c) of the Agricultural Marketing Act of 1946 (7 U.S.C. 
1635j(c)) is amended to read as follows:
    ``(c) Daily Reporting; Barrows and Gilts.--
        ``(1) Prior day report.--
            ``(A) In general.--The corporate officers or officially 
        d... \\ 
 \midrule 
Gold & \tiny Amends the Agricultural Marketing Act of 1946 to extend the provisions of the Livestock Mandatory Price Reporting Act of 1999 through September 30, 2010.

Amends swine price reporting provisions. \\ 
 \midrule 
Model & \small amends the agricultural marketing act of 1946 to terminate the authority of the secretary of agriculture (usda) to: (1) livestock processing plant processing plant slaughter, and (2) slaughtering plant slaughter.

amends the agricultural marketing act of 1946 to: (1) revise minimum reporting requirements; and (2) revise reporting requirements.

amends the agricultural marketing act of 1946 and the agricultural marketing act of 1946 to: (1) revise reporting requirements; and (2) revise reporting requirements. \\ 
 \midrule 
Metrics & Rouge1: 33.66, Rouge2: 18.18, RougeL: 31.68, RougeLsum: 29.7, Summary length (tokens): 105
 \\ 
\\ 
 \midrule 
\multicolumn{2}{l}{Top 75\% example (Sorted by rougeL)}\\ 
 \midrule 
Document & \tiny SECTION 1. SHORT TITLE.

    This Act may be cited as the ``Maritime Administration 
Authorization Act for Fiscal Year 2001''.

SEC. 2. AUTHORIZATION OF APPROPRIATIONS FOR FISCAL YEAR 2001.

    Funds are hereby authorized to be appropriated, as Appropriations 
Acts may provide, for the use of the Department of Transportation for 
the Maritime Administration as follows:
            (1) For expenses necessary for operations and training 
        activities, not to exceed \$80,240,000 for the fiscal year 
        ending September 30, 2001.
            (2) For the costs, as defined in section 502 of the Federal 
        Credit Reform Act of 1990, of guaranteed loans authorized by 
        title XI of the Merchant Marine Act, 1936 (46 U.S.C. App. 1271 
        et seq.), \$50,000,000, to be available until expended. In 
        addition, for administrative expenses related to loan guarantee 
        commitments under title XI of that Act, \$4,179,000.

SEC. 3. AMENDMENTS TO TITLE IX OF THE MERCHANT MARINE ACT, 1936.

    (a) Title IX of the Merchant Marine Act, 1936 (46 U.S.C. App. 101 
et seq.) is amended by adding at the end thereof the following:

``SEC. 910. DOCUMENTATION OF CERTAIN DRY CARGO VESSELS.

    ``(a) In General.--The restrictions of section 901(b)(1) of this 
Act concerning a vessel built in a foreign country shall not apply to a 
newly constructed drybulk or breakbulk vessel over 7,500 deadweight 
tons that has been delivered from a foreign shipyard or contracted for 
construction in a foreign shipyard before the earlier of--
            ``(1) the date that is 1 year after the date of enactment 
        of the Maritime Administration Authorization Act for Fiscal 
        Year 2001; or
            ``(2) the effective date of the OECD Shipbuilding Trade 
        Agreement Act.
    ``(b) Compliance With Certain U.S.-Build Requirements.--A vessel 
timely contracted for or delivered pursuant to this section and 
documented under the laws of the United States shall be deemed to have 
been United-States built for purposes of sections 901(b) and 901b of 
this Act if--
            ``(1) following delivery by a foreign shipyard, the vessel 
        has any additional shipyard work necessary to receive its 
        initial Coast Guard certificate of inspection performed in a 
        United States shipyard;
            ``(2) the vessel is not documented in another country 
        before being documented under the laws of the United States;
            ``(3)... \\ 
 \midrule 
Gold & \tiny (Sec. 3) Amends the Merchant Marine Act, 1936 to declare that certain restrictions concerning a vessel built in a foreign country shall not apply to a newly constructed drybulk or breakbulk vessel over 7,500 deadweight tons that has been delivered from a foreign shipyard or contracted for construction in a foreign shipyard before the earlier of two specified dates. Deems U.S.-built any vessel timely contracted for or delivered and documented under U.S. law, if certain conditions are met.
(Sec. 4) Directs the Secretary of State, in coordination with the Secretary of Transportation, to initiate discussions in all appropriate international forums to establish an international standard for the scrapping of vessels in a safe and environmentally sound manner.
Directs the Secretary of Transportation to develop, and report to specified congressional committees on, a program for the scrapping of obsolete National Defense Reserve Fleet Vessels.
Amends the National Maritime Heritage Act of 1994 to extend, through September 30, 2006, the authority of the Secretary to dispose of certain vessels in the National Defense Reserve Fleet. Requires that such vessels be disposed of in the most cost effective manner to the United States, taking into account the need for disposal, the environment, and safety concerns.
Amends Federal law to authorize the expenditure of funds from the National Defense Sealift Fund for costs related to the scrapping of National Defense Reserve Fleet vessels.
Names vessels in the National Defense Reserve Fleet that may be scrapped in the United States or a foreign country.
(Sec. 5) Requires the Maritime Administration (in its annual report to Congress and its estimated annual budget) to state separately the amount, source, intended use, and nature of any funds (other than funds appropriated to the Administration or to the Secretary for use by the Administration) administered, or subject to oversight, by the Administration.
(Sec. 6) Amends Federal maritime law to authorize the Secretary of Transportation to make a grant to a National Maritime Enhancement Institute for maritime and maritime intermodal research as if the Institute were a university transportation center.
(Sec. 7) Directs the Secretary to study maritime research and technology development, and report the results, including any recommendations, to Congress. Authorizes appropriations.
(Sec. 8) Authorizes the Secretary to convey all right, title, and U.S. interest in the U.S.S. GLACIER (formerly of the National Defense Reserve Fleet) to the Glacier Society, Inc., Bridgeport, Connecticut. \\ 
 \midrule 
Model & \small maritime administration authorization act for fiscal year 2001 - authorizes appropriations for the department of transportation (dot) for fy 2001 for: (1) operations and training activities; (2) training activities; and (3) administrative expenses.amends the merchant marine act, 1936 to make appropriations for fy 2001 through 2001 for the maritime administration.amends the merchant marine act, 1936 to apply certain restrictions concerning a vessel located in a foreign country to a newly constructed dry or breakable vessel over seven,500 feet that has been delivered from a foreign shipyard or contracted for construction in a foreign shipyard before the earlier of: (1) one year after enactment of this act, or (2) the effective date of the international maritime administration act. directs the secretary of state in coordination with the secretary of transportation to initiate discussions in all appropriate international forums in order to establish an international standard for the scrapping of vessels in a safe and environmentally sound manner. directs the secretary of state to initiate discussions in all appropriate international forums to establish an international standard for the scrapping of vessels in a safe and environmentally sound manner. \\ 
 \midrule 
Metrics & Rouge1: 61.19, Rouge2: 41.5, RougeL: 47.76, RougeLsum: 57.21, Summary length (tokens): 222
 \\ 
\\ 
 \midrule 
\multicolumn{2}{l}{Top 95\% example (Sorted by rougeL)}\\ 
 \midrule 
Document & \tiny SECTION 1. SMALL BUSINESS EXPENSING PROVISIONS MADE PERMANENT.

    (a) Increase in Small Business Expensing Made Permanent.--
            (1) In general.--Subsection (b) of section 179 of the 
        Internal Revenue Code of 1986 (relating to limitations) is 
        amended--
                    (A) by striking ``\$25,000 (\$125,000 in the case of 
                taxable years beginning after 2006 and before 2011)'' 
                in paragraph (1) and inserting ``\$500,000'', and
                    (B) by striking ``\$200,000 (\$500,000 in the case of 
                taxable years beginning after 2006 and before 2011)'' 
                in paragraph (2) and inserting ``\$1,000,000''.
            (2) Conforming amendment.--Section 179(b) of such Code is 
        amended by striking paragraph (7).
    (b) Expensing for Computer Software Made Permanent.--Clause (ii) of 
section 179(d)(1)(A) of such Code is amended by striking ``and which is 
placed in service in a taxable year beginning after 2002 and before 
2011,''.
    (c) Inflation Adjustment.--
            (1) So much of subparagraph (A) of section 179(b)(5) of 
        such Code as precedes clause (i) thereof is amended to read as 
        follows:
                    ``(A) In general.--In the case of any taxable year 
                beginning in a calendar year after 2009, the \$500,000 
                and \$1,00,000 amounts in paragraphs (1) and (2) shall 
                each be increased by an amount equal to--''.
            (2) Section 179(b)(5)(A)(ii) of such Code is amended by 
        striking ``2006'' and inserting ``2008''.
    (d) Effective Date.--The amendments made by this section shall 
apply to taxable years ending after the date of the enactment of this 
Act.

SEC. 2. DEDUCTION FOR PURCHASE OF DOMESTICALLY MANUFACTURED 
              AUTOMOBILES.

    (a) In General.--Part VII of subchapter B of chapter 1 of the 
Internal Revenue Code of 1986 (relating to additional itemized 
deductions for individuals) is amended by redesignating section 224 as 
section 225 and by inserting after section 223 the following new 
section:

``SEC. 224. DEDUCTION FOR PURCHASE OF DOMESTICALLY MANUFACTURED 
              AUTOMOBILES.

    ``(a) Allowance of Deduction.--In the case of an individual, there 
shall be allowed as a deduction an amount equal to the cost of any 
qualified automobile placed in service by the taxpayer during the 
taxable year.
    ``(b) Limitation Per Vehicle.--The amount of the ded... \\ 
 \midrule 
Gold & \tiny Amends the Internal Revenue Code to: (1) increase and make permanent the expensing allowance for depreciable business assets; and (2) allow a tax deduction, up to \$10,000, for the purchase of a motor vehicle manufactured in the United States. Terminates such tax deduction after 2010. \\ 
 \midrule 
Model & \small amends the internal revenue code to make permanent: (1) the increased expensing allowance for depreciable business assets; and (2) the tax deduction for the purchase of manufactured manufactured automobiles. \\ 
 \midrule 
Metrics & Rouge1: 72.0, Rouge2: 46.58, RougeL: 64.0, RougeLsum: 64.0, Summary length (tokens): 40
 \\ 

    \bottomrule
    \end{tabular}
}
\caption{Examples of the BillSum dataset using the model billsum-1024-256 base diverse}
\label{tab:examples_billsum-1024-256_base_diverse}
\end{table*}

\begin{table*}[t]
\centering
\resizebox{\textwidth}{!}{
    \begin{tabular}{ l p{21cm} }
    \toprule
\\ 
 \midrule 
\multicolumn{2}{l}{Bottom 5\% example (Sorted by rougeL)}\\ 
 \midrule 
Document & \tiny SECTION 1. SHORT TITLE.

    This Act may be cited as the ``Public Health Equity Act''.

SEC. 2. FINDINGS.

    Congress finds that--
            (1) all communities and individuals are entitled to 
        protection from occupational and other exposure to substances 
        that are hazardous to the public health;
            (2) hazardous substances have had a disproportionate impact 
        on the public health of poor and ethnic minority communities 
        and individuals, resulting in exclusion from participation in, 
        denial of benefits under, and discrimination under, programs 
        and activities receiving Federal financial assistance; and
            (3) each Federal agency has an obligation to ensure that 
        all federally assisted programs and activities that affect 
        human health do not directly or through contractual 
        arrangements use criteria, methods, or practices that cause 
        discrimination on the ground of race, color, or national 
        origin.

SEC. 3. PUBLIC HEALTH EQUITY.

    The Public Health Service Act (42 U.S.C. 201 et seq.) is amended by 
adding at the end thereof the following new title:

                  ``TITLE XXVII--PUBLIC HEALTH EQUITY

``SEC. 2701. DEFINITIONS.

    ``As used in this title:
            ``(1) Activity; program.--The term `program or activity' 
        means any operation of--
                    ``(A)(i) a department, agency, special purpose 
                district, or other instrumentality of a State or of a 
                local government; or
                    ``(ii) the entity of such State or local government 
                that distributes such assistance and each such 
                department or agency (and each other State or local 
                government entity) to which the assistance is extended, 
                in the case of assistance to a State or local 
                government;
                    ``(B)(i) a college, university, or other 
                postsecondary institution, or a public system of higher 
                education; or
                    ``(ii) a local educational agency (as defined in 
                section 198(a)(10) of the Elementary and Secondary 
                Education Act of 1965), system of vocational education, 
                or other school system;
                    ``(C)(i) an entire corporation, partnership, or 
                other private organization, or an entire sole 
              ... \\ 
 \midrule 
Gold & \tiny Public Health Equity Act - Amends the Public Health Service Act to add a new title, title XXVII:  Public Health Equity.  Prohibits, under such title, acts with discriminatory effect as well as intentionally discriminatory acts that result in disproportionate exposure to a covered substance on the basis of race, color, or national origin. Defines a covered substance to include certain hazardous materials, substances, contaminants, or chemicals listed, identified, or defined in specified laws. \\ 
 \midrule 
Model & \small public health equity act - amends the public health service act to require the president to ensure that no person shall be excluded from participation in, be denied the benefits of, or being subject to discrimination under, any program or activity on the ground of race, color, or national origin.

requires the president to ensure that no person shall be excluded from participation in, be denied the benefits of, or being subject to discrimination under, any program or activity on the ground of race, color, or national origin.

requires the secretaries of labor, health and human services, the administrator of the health and human services, the administrator, and any other head of a federal agency with responsibility for providing federal financial assistance to a program or activity to issue regulations implementing such nondiscrimination requirements.

requires such regulations to: (1) declare that no person shall be excluded from participation in, be denied the benefits of, or be subject to discrimination under, any program or activity on the ground of race, color, or national origin; and (2) address actions of programs or activities that result in disproportionate exposure to a covered substance on the basis of race, color, or national origin. \\ 
 \midrule 
Metrics & Rouge1: 28.89, Rouge2: 20.15, RougeL: 22.96, RougeLsum: 26.67, Summary length (tokens): 239
 \\ 
\\ 
 \midrule 
\multicolumn{2}{l}{Bottom 25\% example (Sorted by rougeL)}\\ 
 \midrule 
Document & \tiny SECTION 1. SHORT TITLE; REFERENCES TO TITLE 38, UNITED STATES CODE.

    (a) Short Title.--This Act may be cited as the ``Veterans Programs 
Improvement Act of 2003''.
    (b) References.--Except as otherwise expressly provided, wherever 
in this Act an amendment is expressed in terms of an amendment to a 
section or other provision, the reference shall be considered to be 
made to a section or other provision of title 38, United States Code.

SEC. 2. INCREASE IN RATES OF DISABILITY COMPENSATION AND DEPENDENCY AND 
              INDEMNITY COMPENSATION.

    (a) Rate Adjustment.--The Secretary of Veterans Affairs shall, 
effective on December 1, 2003, increase the dollar amounts in effect 
for the payment of disability compensation and dependency and indemnity 
compensation by the Secretary, as specified in subsection (b).
    (b) Amounts To Be Increased.--The dollar amounts to be increased 
pursuant to subsection (a) are the following:
            (1) Compensation.--Each of the dollar amounts in effect 
        under section 1114.
            (2) Additional compensation for dependents.--Each of the 
        dollar amounts in effect under section 1115(1).
            (3) Clothing allowance.--The dollar amount in effect under 
        section 1162.
            (4) New dic rates.--Each of the dollar amounts in effect 
        under paragraphs (1) and (2) of section 1311(a).
            (5) Old dic rates.--Each of the dollar amounts in effect 
        under section 1311(a)(3).
            (6) Additional dic for surviving spouses with minor 
        children.--The dollar amount in effect under section 1311(b);
            (7) Additional dic for disability.--Each of the dollar 
        amounts in effect under subsections (c) and (d) of section 
        1311.
            (8) DIC for dependent children.--Each of the dollar amounts 
        in effect under sections 1313(a) and 1314.
    (c) Determination of Increase.--
            (1) The increase under subsection (a) shall be made in the 
        dollar amounts specified in subsection (b) as in effect on 
        November 30, 2003.
            (2) Except as provided in paragraph (3), each such amount 
        shall be increased by the same percentage as the percentage by 
        which benefit amounts payable under title II of the Social 
        Security Act (42 U.S.C. 401 et seq.) are increased effective 
        December 1, 2003, as a result of a determination under section 
        215(i) of such Act (42 U.S.C... \\ 
 \midrule 
Gold & \tiny Veterans Programs Improvement Act of 2003 - Directs the Secretary of Veterans Affairs to increase, as of December 1, 2003, the rates of veterans' disability compensation, additional compensation for dependents, the clothing allowance for certain disabled adult children, and dependency and indemnity compensation for surviving spouses and children.

Makes the effective date for the award of death pension the same as that for the award of death compensation or dependency and indemnity compensation.

Excludes lump-sum insurance proceeds from income for purposes of eligibility for veterans' pensions.

Prohibits the payment of veterans' disability compensation for an alcohol- or drug-abuse related disability even if the the alcohol or drug abuse is secondary to a service-connected disability.

Provides alternative beneficiaries for National Service Life Insurance and United States Government Life Insurance proceeds when the first beneficiary does not make a claim.

Provides burial benefit eligibility for a veteran's surviving spouse who remarries following the veteran's death.

Makes permanent the authority for the State cemetery grants program.

Repeals the Department of Veterans Affairs Education Loan program.

Includes self-employment training under the Montgomery GI Bill. \\ 
 \midrule 
Model & \small veterans programs improvement act of 2003 - directs the secretary of veterans affairs, effective on december 1, 2003, to increase the rates of disability and dependency and indemnity compensation (dic) through the department of veterans affairs (va), to: (1) increase the rates of disability compensation and dependency and indemnity compensation; (2) provide for additional compensation for dependents; (3) provide for additional compensation for dependents; (4) exclude lump-sum sales of any life insurance policy or policies on a veteran for purposes of pension benefits; (5) exclude lump-sum sales of any life insurance policy or policies on a veteran for purposes of pension benefits; (6) exclude lump-sum life insurance proceeds from the determinations of annual income for pension purposes; (7) provide for alternative beneficiaries for certain veterans' life insurance policies or policies on a veteran's service-connected disability; and (8) authorize the secretary to approve a program of self-employment on-employment in the department of veterans affairs education loan program.amends the veterans' advisory committee on education to: (1) repeal the requirement that a claimant and the claimant's representative is necessary to complete an application is not received by the secretary within one year from the date of such notification; (2) make permanent the same authority for state cemetery grants program; and (3) authorize the secretary to approve a program of self-employment on-employment in the department of america known as the department of veterans affairs. \\ 
 \midrule 
Metrics & Rouge1: 60.71, Rouge2: 29.79, RougeL: 33.88, RougeLsum: 50.82, Summary length (tokens): 297
 \\ 
\\ 
 \midrule 
\multicolumn{2}{l}{Top 75\% example (Sorted by rougeL)}\\ 
 \midrule 
Document & \tiny SECTION 1. SHORT TITLE.

    This Act may be cited as the ``Cameron Gulbransen Kids and Cars 
Safety Act of 2003''.

SEC. 2. EVALUATION OF DEVICES AND TECHNOLOGY TO REDUCE CHILD INJURY AND 
              DEATH FROM PARKED OR UNATTENDED MOTOR VEHICLES.

    (a) In General.--The Secretary of Transportation shall evaluate--
            (1) devices and technologies intended to reduce the 
        incidence of child injury and child death occurring outside of 
        parked motor vehicles in nontraffic, noncrash events, including 
        backing-over incidents, that are caused by such vehicles, and 
        determining which of those methods is the most effective; and
            (2) currently available technology to prevent injury and 
        death of children left unattended inside of parked motor 
        vehicles, including injury or death due to hyperthermia, power 
        windows, or power sunroofs.
    (b) Report.--The Secretary of Transportation shall submit a report 
on the findings and determinations of the evaluation under this section 
to the Congress by not later than one year after the date of the 
enactment of this Act.
    (c) Completion of Rulemaking Regarding Power Windows.--The 
Secretary of Transportation shall by not later than 6 months after the 
submission of the report under subsection (b) complete any rulemaking 
begun before the date of the enactment of this Act regarding power 
windows and power window switches.

SEC. 3. DATABASE FOR TRACKING THE NUMBER AND TYPES OF INJURIES AND 
              DEATHS IN NONTRAFFIC, NONCRASH EVENTS.

    (a) Establishment.--The Secretary of Transportation shall establish 
a database of (or modify an existing database to include), and collect 
data regarding, the numbers and types of injuries and deaths in 
nontraffic, noncrash events involving motor vehicles.
    (b) Included Information.--The Secretary of Transportation shall 
collect and include in such database the following information:
            (1) The types, makes, models, and model years of motor 
        vehicles involved in nontraffic, noncrash events.
            (2) Whether there was an operator of each motor vehicle in 
        such events.
            (3) The age of each operator of such motor vehicles.
            (4) The age of each individual who suffered injury or death 
        in such events.
            (5) Whether each motor vehicle had technology installed to 
        detect individuals and objects behind it.
            (6... \\ 
 \midrule 
Gold & \tiny Cameron Gulbransen Kids and Cars Safety Act of 2003 - Directs the Secretary of Transportation to: (1) evaluate devices and technologies to reduce child injuries and deaths occurring outside of parked motor vehicles in non-traffic, non-crash events or inside of parked vehicles when children are left unattended; (2) establish a database of, and collect data on, the number and types of injuries and deaths in such events; (3) evaluate technologies for detecting and preventing collisions with individuals and objects behind motor vehicles; (4) prescribe safety standards to require devices for detecting individuals and objects behind motor vehicles; and (5) prescribe safety standards for power windows and power sunroofs, including requirements for child-safe switches and auto reverse technology. \\ 
 \midrule 
Model & \small tamarisk kids and cars safety act of 2003 - directs the secretary of transportation (dot) to evaluate: (1) devices and technologies intended to reduce the incidence of child injury and death occurring inside distant motor vehicles in nontraffic, noncrash events, and determine which are the most effective; and (2) currently available technology to prevent injury and death of children left behind the motor vehicles.

directs the secretary to: (1) establish a database of, and collect data regarding, the number and types of injuries and deaths in nontraffic, noncrash events involving motor vehicles; and (2) prescribe motor vehicle safety standards. \\ 
 \midrule 
Metrics & Rouge1: 63.59, Rouge2: 37.21, RougeL: 50.69, RougeLsum: 49.77, Summary length (tokens): 132
 \\ 
\\ 
 \midrule 
\multicolumn{2}{l}{Top 95\% example (Sorted by rougeL)}\\ 
 \midrule 
Document & \tiny SECTION 1. FINDINGS.

    The Congress finds the following:
            (1) As a Member of Congress from the Tenth Congressional 
        District of Texas, as Majority Leader of the U.S. Senate, Vice-
        President and President of the United States, Lyndon Baines 
        Johnson's accomplishments in the fields of civil rights, 
        education, and economic opportunity rank among the greatest 
        achievements of the past half century.
            (2) As President, Lyndon Johnson proposed, championed, led 
        to passage, and signed into law on August 6, 1965, the Voting 
        Rights Act of 1965, which swept away barriers impeding millions 
        of Americans from meaningful participation in American 
        political life.
            (3) On July 30, 1965, President Johnson signed into law the 
        Social Security Amendments Act of 1965, popularly known as 
        Medicare, which has transformed the delivery of health care in 
        the United States and which, along with Social Security, 
        reduced the rate of poverty among the elderly from 28.5 percent 
        in 1966 to 9.1 percent in 2012.
            (4) On July 2, 1964, President Johnson secured passage and 
        signed into law the most sweeping civil rights legislation 
        since Reconstruction, the Civil Rights Act of 1964, which 
        prohibits discrimination in employment, education, and public 
        accommodations based on race, color, religion, or national 
        origin.
            (5) On November 8, 1965, President Johnson signed into law 
        the Higher Education Act, which provided need-based financial 
        aid to students in the form of scholarships, work-study grants, 
        and loans, and thus made higher education more accessible to 
        populations of persons who were previously unable to attend 
        college because of economic circumstances.
            (6) On October 3, 1965, President Johnson signed into law 
        the Immigration and Naturalization Act of 1965, which 
        transformed the Nation's immigration system by abolishing the 
        racially based quota system that had defined American 
        immigration policy for four decades and replaced it with a 
        policy whose central purpose was family reunification, with a 
        preference for immigrants with specific skill sets.
            (7) According to Robert A. Caro, the preeminent biographer 
        of Lyndon Baines Johnson, with the s... \\ 
 \midrule 
Gold & \tiny This bill directs the Speaker of the House and the President pro tempore of the Senate to arrange for the posthumous award of a Congressional Gold Medal to Lyndon Baines Johnson in recognition of his contributions to the nation, including passage of the Voting Rights Act of 1965, the Social Security Amendments Act (Medicare) of 1965, the Civil Rights Act of 1964, the Higher Education Act of 1965, and the Immigration and Naturalization Act of 1965. Requires such medal to be given to the Lyndon Baines Johnson Library and Museum following its award, where it will be available for display and research.  \\ 
 \midrule 
Model & \small this bill directs the speaker of the house of representatives and the president pro tempore of the senate to arrange for the posthumous award, on behalf of congress, of a gold medal to lyrics to lyrics in recognition of his contributions to the nation, including recognition of his contributions to the nation, including recognition of the landmark voting rights act of 1965, the civil rights act of 1964, the higher education act of 1965, and the immigration and naturalization act of 1965. \\ 
 \midrule 
Metrics & Rouge1: 72.83, Rouge2: 62.64, RougeL: 68.48, RougeLsum: 68.48, Summary length (tokens): 97
 \\ 

    \bottomrule
    \end{tabular}
}
\caption{Examples of the BillSum dataset using the model billsum-4096-1024 base diverse}
\label{tab:examples_billsum-4096-1024_base_diverse}
\end{table*}

\begin{table*}[t]
\centering
\resizebox{\textwidth}{!}{
    \begin{tabular}{ l p{21cm} }
    \toprule
\\ 
 \midrule 
\multicolumn{2}{l}{Bottom 5\% example (Sorted by rougeL)}\\ 
 \midrule 
Document & \tiny this study is an extension of a report on patients with type 1 diabetes at children 's hospital of new orleans ( 14 ) and was approved by the institutional review board at louisiana state university health sciences center , new orleans , louisiana . 
 glucose data were downloaded from patient meters at each clinic visit . meter model and sampling protocols varied by patient preference and insurance provider . 
 an average of three glucose measurements per day were recorded in a study using a similar self - monitoring protocol ( 7 ) . 
 a1c was measured by national glycohemoglobin standardization program ( ngsp)-approved immunoassays ( 15 ) at the children 's hospital ( 184 patients ) or by commercial laboratories that presumably also used ngsp - approved methods ( 18 patients , including 4 low- , 7 moderate- , and 7 high - hgi subjects ) . 
 a population regression equation \{ a1c ( \% ) = [ 0.021  mbg ( mg / dl ) ] + 4.3 , r = 0.57 \} was derived using mean mbg and mean a1c from 202 patients collected at 1,612 clinic visits as described elsewhere ( 14 ) . 
 the same data were used to calculate hemoglobin glycation index ( hgi ) and to divide patients into low- , moderate- , and high - hgi groups . predicted a1c values were calculated at each clinic visit by inserting mbg into the regression equation . 
 hgi values were calculated by subtracting predicted a1c from observed a1c measured at the same clinic visit . 
 patients were divided into low- , moderate- , and high - hgi groups based on mean hgi tertile ( 33\% ) rank ( low hgi , \textless{} 0.41 , n = 67 ; moderate hgi , 0.41 to 0.26 , n = 68 ; high hgi , \textgreater{} 0.26 , n = 67 ) . 
 eag was calculated by inserting observed a1c into the adag linear regression equation ( eag [ mg / dl ] = [ 28.7  a1c ( \% ) ]  46.7 , r = 0.92 ) ( 1 ) . 
 a mean blood glucose index ( mbgi ) that quantifies the difference between mbg and eag was calculated by subtracting observed mbg from eag . 
 descriptive statistics and linear regression analyses were generated using graphpad prism v. 4.03 ( graphpad software , san diego , ca ) . 
 glucose data were downloaded from patient meters at each clinic visit . meter model and sampling protocols varied by patient preference and insurance provider . 
 an average of three glucose measurements per day were recorded in a study using a similar self - monitoring protocol ( 7 ) . 
 a1c was measured by national glycohemoglobin standardization program ( ngsp)-approved immunoassays ( 15 ) at the children 's ho... \\ 
 \midrule 
Gold & \tiny objectivethe a1c - derived average glucose study recommended reporting a1c in estimated average glucose ( eag ) equivalents . 
 we compared eag with self - monitored mean blood glucose ( mbg ) to determine whether eag is systematically biased due to biological variation in the relationship between mbg and a1c.research design and methodsmbg and a1c were recorded from charts of 202 pediatric type 1 diabetic patients at 1,612 clinic visits 
 . patients were divided into groups with low , moderate , or high a1c bias based on a hemoglobin glycation index ( hgi).resultsthe mean  sd values for mbg versus eag were as follows : total population , 194  34 vs. 196  36 mg / dl ; low - hgi group , 186  31 vs. 163  20 mg / dl ; moderate - hgi group , 195  28 vs. 193  19 mg / dl ; and high - hgi group , 199  42 vs. 230  31 mg / dl.conclusionseag underestimated mbg in low hgi patients and overestimated mbg in high hgi patients . 
 disagreement between eag and mbg downloaded from patient glucose meters will cause confusion if eag is implemented for clinical use . \\ 
 \midrule 
Model & \small objectiveto investigate the relationship between hemoglobin glycated hemoglobin ( hba1c ), hemoglobin glycated hemoglobin ( hba1c ), and hemoglobin glycated hemoglobin ( hba1c ) in children with type 1 diabetes.research design and methodsa1c was measured by national glycohemoglobin standardization program ( ngsp)-approved immunoassays at the children's hospital of new orleans. 
 hba1c was measured by national glycohemoglobin standardization program ( ngsp)-approved immunoassays at the children's hospital of new orleans. 
 hba1c was measured by national glycohemoglobin standardization program ( ngsp)-approved immunoassays at the children's hospital of new orleans. 
 a population regression equation [ a1c ( \% ) = [ 0.021  mbg ( mg / dl ] + 4.3, r = 0.92 ) + 4.3, r = 0.58 ] was derived using mean hba1c ( \% ). 
 a population regression equation [ a1c ( \% ) = [ 0.021  mbg ( mg / dl ] + 4.3, r = 0.92 ) + 4.3, r = 0.92 ] was \\ 
 \midrule 
Metrics & Rouge1: 19.4, Rouge2: 6.06, RougeL: 10.03, RougeLsum: 13.38, Summary length (tokens): 256
 \\ 
\\ 
 \midrule 
\multicolumn{2}{l}{Bottom 25\% example (Sorted by rougeL)}\\ 
 \midrule 
Document & \tiny the principal aim of this laboratory is the synthesis of conjugated unsaturated ketones as candidate antineoplastic agents . 
 these compounds interact with thiols but in general , they have little or no affinity for amino and hydroxyl groups which are found in nucleic acids ( 1 - 3 ) . 
 hence thiol alkylators may not have the genotoxic properties associated with a number of contemporary anticancer drugs ( 4 ) . 
 however after an initial chemical insult , certain neoplasms are more vulnerable to a subsequent cytotoxic effect than various non - malignant cells ( 5 , 6 ) . 
 hence by mounting the 1,5-diaryl-3-oxo-1,4-pentadienyl pharmacophore [ ar - c = c - c(o)-c = c - ar ] on heterocyclic and cycloaliphatic scaffolds , two sequential interactions with cellular thiols can take place which may be more detrimental to tumours than normal tissues . 
 such considerations led to the development of 3,5-bis(benzylidene)-4-piperidones 1a - d which demonstrated potent cytotoxic properties with the ic50 values in the low micromolar range against human molt 4/c8 and cem t - lymphocytes as well as murine l1210 lymphocytic leukemia cells ( 7 , 8) . 
 the hypothesis of sequential cytotoxicity was advanced that the 1,5-diaryl-3-oxo-1,4-pentadienyl group interacts at a primary binding site and a side chain on the piperidine nitrogen may align at an auxiliary binding site which could enhance cytotoxic potencies . in order to evaluate this hypothesis , a novel series of n - aroyl-3,5-bis(benzylidene)-4-piperidone derivatives 2 - 7 were synthesized ( 8) . in these compounds , the side chain contains atoms and groups capable of forming hydrogen and van der waals bonds and also an ionic bond in series 3 - 6 which could bind at an auxiliary binding site . when the potencies of the compounds in series 3 - 6 were compared with that of the analogs 1a - d which have the same aryl substituent , the ic50 values of the amides 3 - 6 were lower in 48\% of the comparisons made while equipotency was noted in 35\% of the cases ( 8) . 
 the proposed primary and auxiliary binding sites of cytotoxic n - aroyl-3,5-bis(benzylidene)-4-piperidone analogs   the objectives of the present investigation are two folds . 
 first , to gain some insight into the physicochemical properties of the 3,5-bis(benzylidene)-4-piperidone derivatives 1 - 7 that influence cytotoxic potencies , and second , to derive predictive 2d - qsar models that will be helpful to design new analogs with improved bioactivity . 
 t... \\ 
 \midrule 
Gold & \tiny 1-[4-(2-alkylaminoethoxy ) phenylcarbonyl]-3,5-bis(arylidene)-4-piperidones are a novel class of potent cytotoxic agents . 
 these compounds demonstrate low micromolar to submicromolar ic50 values against human molt 4/c8 and cem t - lymphocytes and murine leukemia l1210 cells . in this study , 
 a comparative qsar investigation was performed on a series of 3,5-bis ( arylidene)-4-piperidones using different chemometric tools to develop the best predictive models for further development of analogs with improved cytotoxicity . 
 all the qsar models were validated by internal validation tests . 
 the qsar models obtained by ga - pls method were considered the best as compared to mlr method . the best qsar model obtained by ga - pls analysis on l1210 , cem and molt4/c8 demonstrated good predictively with r2pred values ranging from 0.94 - 0.80 . molecular density , 
 topological ( x2a ) and geometrical indices of the molecules were found to be the most important factors for determining cytotoxic properties . \\ 
 \midrule 
Model & \small in order to evaluate the physicochemical properties of the 3,5-bis(benzylidene)-4-piperidone derivatives 1 - 7 and 2 - 7 were synthesized. 
 the compounds 1 - 7 showed potent cytotoxic property with ic50 values of 3 - 6. 
 the compounds 1 - 7 showed potent cytotoxic property with ic50 values of 3 - 6. 
 the compounds 1 - 7 showed potent cytotoxic property with ic50 values of 3 - 6. 
 the compounds 1 - 7 showed potent cytotoxic property with ic50 values of 3 - 6. 
 the compounds 1 - 7 showed potent cytotoxic property with ic50 values of 3 - 6. 
 the compounds 1 - 7 showed potent cytotoxic property with ic50 values of 3 - 6. 
 the compounds 1 - 7 showed potent cytotoxic property with ic50 values of 3 - 6. 
 the compounds 1 - 7 showed potent cytotoxic property with ic50 values of 3 - 6. 
 the compounds 1 - 7 showed potent cytotoxic property with ic50 values of 3 - 6. 
 the compounds 1 - 7 showed potent cytotoxic property with ic50 values of 3 - 6. \\ 
 \midrule 
Metrics & Rouge1: 21.38, Rouge2: 5.06, RougeL: 16.35, RougeLsum: 16.35, Summary length (tokens): 253
 \\ 
\\ 
 \midrule 
\multicolumn{2}{l}{Top 75\% example (Sorted by rougeL)}\\ 
 \midrule 
Document & \tiny educational evaluation ( ee ) is a formal process performed to evaluate the quality of effectiveness and/or value of a program , process , goal or curriculum.12 it deals with data collection and assessment of the progress of academic programs.34 by considering some principles related to educational measurements and data collection , ee may result in a better understanding of such programs.57 during the past thirty years , theorists have presented numerous methods of evaluation . 
 worthen and sanders2 mentioned that more than 50 different evaluation approaches has been developed in recent decades . among these 
 , methods based on internal criteria are known as the ones that can interpret the scientific , educational , and therapeutic authenticity of different educational groups.48 this is greatly welcomed by the academic community and is widely spread to all universities in the world . 
 that is because this method provided a scientific , appropriate , precise , timely , and valid basis regarding the interpretation of decision making system quality and programming for its promotion and development.3 such a method was successfully carried out in four medical education groups at supervisory and expansion of medical education council secretariat of ministry of health , treatment and medical education of iran in 1995.8 ee has its most effect , value , and results when it can provide needed information to individuals which are directly related , as well as those who may be benefited from its results.346 educating dental professionals consists of theoretical and practical ( clinical , paraclinical , and laboratory ) courses , differing in duration , and educational curriculum among different countries . 
 it might vary from 4 years ( e.g. , in india , turkey , and russia ) to 6 years or more ( e.g. , in iran consisting of 2 years of only basic medical sciences and 4 + years of dentistry courses ) . due to numerous practical educational units in dentistry education and with regard to expensive but very critical protocols for infection control , a great deal of resources 
 is consumed in governmental universities of iran over training every general practitioner with a degree of  doctorate of dental surgery  . 
 on the other hand , the quality of dental services plays an important role in public health . considering these issues , 
 dentistry education needs to be cost - effective in terms of optimizing its quality . to date , the pace of advancements in dentistry... \\ 
 \midrule 
Gold & \tiny background : educational evaluation is a process which deals with data collection and assessment of academic activities progress . in this research , educational evaluation of dentistry school of tehran university of medical sciences , which trains students in undergraduate and residency courses , was studied.methods:this descriptive study was done with a model of educational evaluation in ten steps and 13 fields including purposes and mission objectives , management and organization , academic board members , students , human resources and support , educational , research , health and treatment spaces , educational , diagnostic , research and laboratory tools , educational , research , health and treatment programs and courses , process of teaching and learning , evaluation and assessment , alumni , and patients satisfaction . 
 data were collected using observation , interviews , questionnaires , and checklists.results:results of the study were mainly qualitative and in some cases quantitative , based on defined optimal situation . 
 the total mean of qualitative results of educational evaluation of dentistry school in all 13 fields was 55.98\% which is relatively desirable . in the case of quantitative ones , 
 results of some fields such as treatment quality of patients and education and learning of the students were relatively desirable ( 61.32\% and 60.16\% respectively).conclusion : according to the results , educational goals and missions , educational and research facilities and spaces which were identified as the weakest areas need to be considered and paid more serious attention . \\ 
 \midrule 
Model & \small background : educational evaluation ( ee ) is a formal process performed to evaluate the quality of effectiveness and/or value of a program, process, goal or curriculum. 
 it deals with data collection and assessment of the progress of academic programs. 
 the objective of this descriptive cross - sectional study was to assess the quality of education provided at dental school of tehran university of medical sciences.materials and methods : in this descriptive cross - sectional study, the efficacy of provided educational programs provided at dental school of tehran university of medical sciences educational groups in 2010 through this descriptive cross - sectional study was assessed in 13 fields. 
 the data were analyzed using spss version 20.results:the overall quality of education provided in 13 fields. 
 the most important quality of education was the quality of education. 
 the most important quality of education was the quality of education. 
 the most important quality of education was the quality of education. 
 the most important quality of education was the quality of education. 
 the most important quality of education was the quality of education. 
 the most important quality of education was the quality of education. 
 the most important quality of education was the quality of education. 
 the most important quality of education was the quality of education. \\ 
 \midrule 
Metrics & Rouge1: 39.32, Rouge2: 12.68, RougeL: 27.67, RougeLsum: 34.47, Summary length (tokens): 256
 \\ 
\\ 
 \midrule 
\multicolumn{2}{l}{Top 95\% example (Sorted by rougeL)}\\ 
 \midrule 
Document & \tiny the study population included patients over 18  years old who had an initial cabg or combined cabg and open chest aortic valve replacement ( avr ) , from april 1 , 1998 to october 31 , 2011 in ontario . 
 the date of first cardiac surgery was the index date , and eligible patients were followed for 1  year with respect to major outcomes , and 5  years for mortality . 
 preoperative data were included for 1  year prior to surgery , and outcomes for 1  year postoperatively . 
 patients for whom sex , age , height , weight were missing , and patients living outside of ontario or of unknown residence were excluded . cardiac care network of ontario ( ccn ) 
 data were used to identify baseline characteristics such as cardiac ejection fraction , number of grafts bypassed , prior myocardial infarction ( mi ) , emergency or elective surgery , and other co - morbidities . 
 ccn data and the following datasets were combined from ices using deterministic linkage by unique ices key number identifiers : ontario health insurance plan , canadian institute of health information ( cihi ) discharge abstract database , national ambulatory care reporting system , same day surgery , and the registered persons database . 
 patients who had undergone either isolated cabg or combined cabg / avr were selected from the cihi discharge abstract database . 
 data for which other cardiac procedures had been performed during the same admission were excluded ( eg , percutaneous coronary intervention or other valve procedures ) . 
 bmi was calculated as weight ( kg)/height ( m ) , and patients were divided into groups : underweight ( bmi \textless{} 20  kg / m ) , normal weight ( bmi 20.0 to 24.9  kg / m ) , overweight ( bmi 25.0 to 29.9  kg / m ) , obese ( bmi 30.0 to 34.9  kg / m ) , and morbidly obese ( bmi \textgreater{} 34.9 
 kg / m ) , closely based on world health organization ( who ) and health canada guidelines.1214 the following comorbidities were assessed for presence within 1  year prior to index date : diabetes , smoking history ( current or ever smoked ) , peripheral vascular disease ( pvd ) , chronic obstructive pulmonary disease ( copd ) , dialysis within 1  year prior to surgery , cerebrovascular disease ( cvd ) , congestive heart failure ( chf , from cihi ) , hypertension , elective or emergent surgery , creatinine . within 30  days 
 prior to surgery , the following cardiac characteristics were captured : ejection fraction , prior mi , left main coronary disease , and previous cabg at date o... \\ 
 \midrule 
Gold & \tiny backgroundthe  obesity paradox  reflects an observed relationship between obesity and decreased morbidity and mortality , suggesting improved health outcomes for obese individuals . 
 studies examining the relationship between high body mass index ( bmi ) and adverse outcomes after cardiac surgery have reported conflicting results.methods and resultsthe study population ( n=78  762 ) was comprised of adult patients who had undergone first - time coronary artery bypass ( cabg ) or combined cabg / aortic valve replacement ( avr ) surgery from april 1 , 1998 to october 31 , 2011 in ontario ( data from the institute for clinical evaluative sciences ) . 
 perioperative outcomes and 5-year mortality among pre - defined bmi ( kg / m2 ) categories ( underweight \textless{} 20 , normal weight 20 to 24.9 , overweight 25 to 29.9 , obese 30 to 34.9 , morbidly obese \textgreater{} 34.9 ) were compared using bivariate analyses and cox multivariate regression analysis to investigate multiple confounders on the relationship between bmi and adverse outcomes . 
 a reverse j - shaped curve was found between bmi and mortality with their respective hazard ratios . 
 independent of confounding variables , 30-day , 1-year , and 5-year survival rates were highest for the obese group of patients ( 99.1\% [ 95\% confidence interval \{ ci \} , 98.9 to 99.2 ] , 97.6\% [ 95\% ci , 97.3 to 97.8 ] , and 90.0\% [ 95\% ci , 89.5 to 90.5 ] , respectively ) , and perioperative complications lowest . 
 underweight and morbidly obese patients had higher mortality and incidence of adverse outcomes.conclusionsoverweight and obese patients had lower mortality and adverse perioperative outcomes after cardiac surgery compared with normal weight , underweight , and morbidly obese patients . 
 the  obesity paradox  was confirmed for overweight and moderately obese patients . 
 this may impact health resource planning , shifting the focus to morbidly obese and underweight patients prior to , during , and after cardiac surgery . \\ 
 \midrule 
Model & \small backgroundthe aim of this study was to investigate the relationship between body mass index ( bmi ) and mortality in patients undergoing coronary artery bypass grafting ( cabg ) or open chest aortic valve replacement ( avr).methods and resultswe identified 1  year patients undergoing primary cabg or combined cabg and open chest aortic valve replacement ( avr ) from april 1, 1998 to october 31, 2011. 
 patients were divided into groups : underweight ( bmi \textless{} 20  kg / m2 ), normal weight ( bmi 25.0 to 29.9  kg / m2 ), overweight ( bmi 25.0 to 29.9  kg / m2 ), obese ( bmi 30.0 to 29.9  kg / m2 ), and morbidly obese ( bmi \textgreater{} 34.9  kg / m2 ). 
 patients were followed for 1  year with respect to major outcomes, and 5  years for mortality. 
 cox proportional hazards regression analysis was used to investigate multiple confounders on the relationship between bmi and mortality, providing hazard ratios and 95\% confidence intervals ( ci ). 
 patients with bmi \textless{} 20  kg / m2 were more likely to die during the 5  years of follow - up ( hazard ratio [ hr ] \\ 
 \midrule 
Metrics & Rouge1: 56.82, Rouge2: 26.29, RougeL: 40.91, RougeLsum: 50.0, Summary length (tokens): 256
 \\ 

    \bottomrule
    \end{tabular}
}
\caption{Examples of the PubMed dataset using the model pubmed-4096-512 small diverse}
\label{tab:examples_pubmed-4096-512_small_diverse}
\end{table*}

\begin{table*}[t]
\centering
\resizebox{\textwidth}{!}{
    \begin{tabular}{ l p{21cm} }
    \toprule
\\ 
 \midrule 
\multicolumn{2}{l}{Bottom 5\% example (Sorted by rougeL)}\\ 
 \midrule 
Document & \tiny in august , 4 months before presentation , a 35-year - old white woman of scots and english descent developed reddish urine for several days followed by eruption of vesicles and blisters on the dorsal surfaces of her hands and fingers , the sides of her nose , and her upper anterior chest , knees , and legs . 
 she worked as a landscaping contractor and noticed that lesions occurred on areas exposed to sunlight , but application of sunscreen neither diminished the rate at which new lesions appeared , nor promoted healing of older lesions . 
 her skin was fragile in areas of the lesions and the lesions healed slowly , often with scarring . 
 she also developed dark brown pigmentation and the growth of fine black hair over her cheeks . 
 she consumed three glasses of wine each week and had smoked electronic cigarettes for approximately 6 months , having changed from tobacco cigarettes . 
 she had donated three units of blood for transfusion , but none in several years . 
 she had no menses in the 12 months before presentation due to the effects of a contraceptive vaginal ring ( nuvaring ; etonogestrel / ethinyl estradiol ) . 
 a dermatologist performed a punch biopsy of two skin lesions on her left forefinger and referred her for hematology evaluation and treatment . 
 physical examination confirmed the presence of new vesicles and bullae with erythematous bases , some as large as 1 cm in diameter , and older lesions in various stages of erosion , resolution , and scarring in the anatomical distribution described above . 
 lesions were most prominent on the dorsal surfaces of the hands and fingers ( fig . 
 milia were scattered over areas affected with bullous lesions and were especially prominent on skin overlying finger joints . 
 ( a ) before therapy , there were vesicles , bullae with erythematous bases , older lesions in various stages of erosion , resolution , scarring , and punctate milia , all typical of pct . 
 ( b ) after iron depletion with phlebotomy , there were no active skin lesions , although hyperpigmentation , scarring , and milia ( especially over dorsal aspects of second and third interphalangeal joints ) persisted . 
 punch biopsy specimens of skin were deposited in immunofluorescence transport medium , flash frozen , and cut for manual immunofluorescence staining . 
 the sections were probed with fluorescein - labeled anti - human antibodies specific for igg , iga , igm , c3 , c5b-9 , and fibrinogen . 
 complete blood count ( including ... \\ 
 \midrule 
Gold & \tiny porphyria cutanea tarda ( pct ) is characterized by decreased uroporphyrinogen decarboxylase activity in hepatocytes , uroporphyrin i and heptacarboxyl porphyrin iii accumulation , photosensitivity dermatitis , and increased storage iron . in women , estrogen therapy , including oral contraceptives , postmenopausal hormone replacement , and tamoxifen for breast cancer treatment , is a risk factor for pct . 
 we report the case of a woman who presented with pct , hfe c282y homozygosity , and hepatic iron overload and was using a contraceptive vaginal ring containing ethinyl estradiol , an estrogen . 
 we discuss this case in the context of characteristics of other persons with pct , including common hfe mutations , iron overload , and estrogen exposure . \\ 
 \midrule 
Model & \small a 35-year - old white woman of scots and english descent developed reddish urine for several days followed by eruption of vesicles and blisters on the dorsal surfaces of her hands and fingers, the sides of her nose, and her upper anterior chest, knees, and legs. 
 skin biopsy specimens revealed linear glassy iga and igg deposition along the epidermal basement membrane zone and superficial dermal blood vessels, forming a  doughnut pattern. 
 the patient was treated with a levonorgestrel - releasing intrauterine contraceptive system ( skyla ). 
 eight months after presentation, she underwent implantation of a levonorgestrel - releasing intrauterine contraceptive system ( skyla ). 
 eight months after presentation, she underwent implantation of a levonorgestrel - releasing intrauterine contraceptive system ( skyla ). \\ 
 \midrule 
Metrics & Rouge1: 20.56, Rouge2: 0.94, RougeL: 11.22, RougeLsum: 17.76, Summary length (tokens): 176
 \\ 
\\ 
 \midrule 
\multicolumn{2}{l}{Bottom 25\% example (Sorted by rougeL)}\\ 
 \midrule 
Document & \tiny this was a multicenter , community - based , retrospective observational study of patients with pns , ranging from 8 to 20 mm in diameter , presenting to 18 geographically representative outpatient pulmonary clinics across the united states . 
 the study was approved at 15 sites by a central institutional review board and at three sites by local institutional review board approval . 
 four hundred forty sites were identified based on investigator databases and claims data from a large insurance carrier whose coverage population was representative of the overall us population . of these 
 , 77 sites expressed interest in participating , and 48 sites went on to sign confidentiality agreements . 
 of these , 17 did not request additional information , leaving 31 sites undergoing qualification review . 
 eighteen outpatient pulmonary clinics were chosen to participate based on the following criteria : ( 1 ) management of patients with pns , ( 2 ) availability of medical records , and ( 3 ) ability to perform data abstraction . 
 in addition , investigators targeted enrollment of geographically diverse patients to limit the potential bias associated with differences in practice patterns and to account for variation in disease prevalence ( eg , endemic mycoses ) that could alter management decisions . 
 patients were identified by querying databases ( eg , billing and scheduling systems ) using five international classification of diseases , ninth revision , clinical modification codes for pn ( 793.1 , 786.6 , 518.89 , 519.8 , 519.9 ) to ensure homogeneity in patient identification and inclusion . 
 manual chart abstraction was then used to identify those who met the criteria . to minimize selection bias , the sites were not permitted to use additional codes during database query to identify patients . to ensure a systematic sample , 
 inclusion criteria included age  40 years and  89 years at the time of nodule finding , presentation to a pulmonologist , nodule size 8 to 20 mm , and definitive diagnosis ascertained by tissue diagnosis or radiographic follow - up for 2 years . 
 exclusion criteria included chest ct scan performed \textgreater{} 60 days prior to the initial visit , prior diagnosis of any cancer within 2 years of nodule detection , or incomplete chart data . 
 patients were categorized into three groups by the most invasive procedure performed during management , as follows : surveillance ( serial imaging ) , biopsy ( ct scan - guided transthoracic needle aspi... \\ 
 \midrule 
Gold & \tiny background : pulmonary nodules ( pns ) are a common reason for referral to pulmonologists . 
 the majority of data for the evaluation and management of pns is derived from studies performed in academic medical centers . 
 little is known about the prevalence and diagnosis of pns , the use of diagnostic testing , or the management of pns by community pulmonologists.methods:this multicenter observational record review evaluated 377 patients aged 40 to 89 years referred to 18 geographically diverse community pulmonary practices for intermediate pns ( 8 - 20 mm ) . study measures included the prevalence of malignancy , procedure / test use , and nodule pretest probability of malignancy as calculated by two previously validated models . 
 the relationship between calculated pretest probability and management decisions was evaluated.results:the prevalence of malignancy was 25\% ( n = 94 ) . nearly one - half of the patients ( 46\% , n = 175 ) had surveillance alone . 
 biopsy was performed on 125 patients ( 33.2\% ) . 
 a total of 77 patients ( 20.4\% ) underwent surgery , of whom 35\% ( n = 27 ) had benign disease . 
 pet scan was used in 141 patients ( 37\% ) . 
 the false - positive rate for pet scan was 39\% ( 95\% ci , 27.1\%-52.1\% ) . 
 pretest probability of malignancy calculations showed that 9.5\% ( n = 36 ) were at a low risk , 79.6\% ( n = 300 ) were at a moderate risk , and 10.8\% ( n = 41 ) were at a high risk of malignancy . 
 the rate of surgical resection was similar among the three groups ( 17\% , 21\% , 17\% , respectively ; p = .69).conclusions : a substantial fraction of intermediate - sized nodules referred to pulmonologists ultimately prove to be lung cancer . despite advances in imaging and nonsurgical biopsy techniques , invasive sampling of low - risk nodules and surgical resection of benign nodules remain common , suggesting a lack of adherence to guidelines for the management of pns . \\ 
 \midrule 
Model & \small objectivethe aim of this study was to document the prevalence of cancer ( 25\% ) in patients with intermediate - sized pulmonary nodules referred to community pulmonologists.materials and methodsthis was a multicenter, community - based, observational study of patients with intermediate - sized pulmonary nodule who were referred to pulmonologists. 
 patients were categorized into three groups by the most invasive procedure performed during management, as follows : surveillance ( serial imaging ), biopsy ( ct - guided transthoracic needle aspiration [ ttna ] or bronchoscopy ), or surgery ( including mediastinoscopy, video - assisted thorascopic surgery, and/or thoracotomy ). 
 patients were categorized into three groups by the most invasive procedure performed during management, as follows : surveillance ( serial imaging ), biopsy ( ct - guided transthoracic needle aspiration [ ttna ] or bronchoscopy ), or surgery ( including mediastinoscopy, video - assisted thorascopic surgery, and/or thoracotomy ). 
 patients were categorized into three groups by the most invasive procedure performed during management, as follows : surveillance ( serial imaging ), biopsy ( ct - guided transthoracic needle aspiration [ ttna ] or bronchoscopy ), or surgery ( including mediastinoscopy, video - assisted thorascopic surgery, and/or thoracotomy ). 
 multivariate logistic regression was performed to identify factors associated with the use of an invasive diagnostic procedure.resultsof the 377 patients included, 283 ( 75\% ) had a nodule that was benign, and 94 ( 25\% ) had a malignant nodule. 
 the overall accuracy of pet scanning was 74\%, with a false - positive ( fp ) rate of 39\% and a false - negative ( fn ) rate of 9\%. 
 the overall accuracy of pet scanning was 74\%, with a false - positive ( fp ) rate of 39\% and a false - negative ( fn ) rate of 9\%. 
 nodules measuring \textgreater{} 11 to 15 mm ( n = 48 ) had fn and fp rates of 9\% and 36\%, respectively.conclusionsthe prevalence of cancer in patients with intermediate - sized nodules was 25\%. 
 the rate of surgical resection for benign disease varied from 9\% to 23\% in screening trials and surgical series. \\ 
 \midrule 
Metrics & Rouge1: 45.58, Rouge2: 9.56, RougeL: 18.37, RougeLsum: 38.1, Summary length (tokens): 470
 \\ 
\\ 
 \midrule 
\multicolumn{2}{l}{Top 75\% example (Sorted by rougeL)}\\ 
 \midrule 
Document & \tiny a total of 1,217 dead birds were shipped at 4c to the tropical medicine institute " pedro kouri " and identified by ornithology experts . 
 brain , heart , and kidneys were removed and tested for wnv by using reverse transcription  polymerase chain reaction ( rt - pcr ) ( 12 ) . 
 briefly , rna was extracted by using the qiamp viral rna kit ( qiagen , inc . , 
 valencia , ca , usa ) . primers wn212 ( 5-ttgtgttggctctcttggcgttctt-3  ) and wn619c ( 5-cagccgacagcactggacattcata-3  ) were used to detect viral rna . a second rt - pcr with primers wn9483 ( 5-cacctacgccctaaacactttcacc-3  ) and wn9794 ( 5-ggaacctgctgccaatcataccatc-3  ) was performed on the same rna preparation . 
 serum specimens from horses in havana and havana province were tested for antibodies to wnv by using a competitive enzyme - linked immunosorbent assay ( elisa ) with monoclonal antibodies 3.1112 g and 6b6c-1 as described by blitvich et al . 
 we tested 210 serum specimens from horses collected as part of an infectious anemia study . 
 the immunoglobulin m ( igm ) test was not performed because horses were never suspected of having wnv and did not have any history of suspected viral encephalitis or other illness or symptoms . 
 an inhibition value \textgreater{} 30\% was used as the diagnostic criterion to identify flavivirus antibody ( table 1 ) . 
 * wnv , west nile virus ; elisa , enzyme - linked immunosorbent assay ; prnt , plaque reduction neutralization test ; slev , saint louis encephalitis virus . the cuban health ministry and medical services division conducted surveillance for encephalitis of unknown origin in patients \textgreater{} 30 years of age . 
 serum and cerebrospinal fluid specimens were shipped at 4c to the tropical medicine institute " pedro kouri . " 
 human sera were screened for wnv igm and igg by using commercial igm and igg elisa kits ( focus technologies , cypress , ca , usa ) according to manufacturer 's instructions . 
 hemagglutination - inhibition ( hi ) tests were also undertaken with wnv and saint louis encephalitis virus ( slev ) antigen ( 14 ) . 
 reactive serum samples were further tested by a plaque reduction neutralization test ( prnt ) with wnv ( ny99 , ontario , canada , 2001 isolate ) , slev ( parton strain , american type culture collection catalog no . 
 vr-1265 ) , and dengue virus ( dengue 2 , ng - c strain ) . 
 prnt was performed to confirm wnv - specific antibody and was carried out as described previously ( 15 ) by using a neutral red double - overlay procedure . 
 ho... \\ 
 \midrule 
Gold & \tiny a surveillance system to detect west nile virus ( wnv ) was established in cuba in 2002 . 
 wnv infection was confirmed by serologic assays in 4 asymptomatic horses and 3 humans with encephalitis in 2003 and 2004 . 
 these results are the first reported evidence of wnv activity in cuba . \\ 
 \midrule 
Model & \small we report the first evidence of antibodies to west nile virus ( wnv ) in horses and humans in cuba. 
 antibodies were detected by plaque reduction neutralization test in 10 of 1,217 dead birds tested. 
 these findings provide evidence that wnv and slev may co - circulate in cuba. \\ 
 \midrule 
Metrics & Rouge1: 51.06, Rouge2: 17.39, RougeL: 31.91, RougeLsum: 46.81, Summary length (tokens): 64
 \\ 
\\ 
 \midrule 
\multicolumn{2}{l}{Top 95\% example (Sorted by rougeL)}\\ 
 \midrule 
Document & \tiny intra - articular injections of corticosteroids have been used for several decades in the management of inflammatory and degenerative joint conditions when first - line conservative therapies such as rest , ice , and anti - inflammatory medications fail to provide adequate symptom relief . based in part on this long history of successful utilization coupled with the findings of several randomized controlled trials , consensus statements and meta - analyses 
 have concluded that intra - articular corticosteroid injections provide short - term patient benefit and clinical efficacy for chronic knee pain.13 more recently , various injectable hyaluronic acid agents have become commercially available and have enjoyed widespread clinical acceptance as an effective treatment for knee osteoarthritis . 
 these agents are indicated for the treatment of the pain associated with osteoarthritis of the knee in patients who have failed to respond adequately to conservative nonpharmacologic therapy and simple analgesics , eg , acetaminophen . 
 traditionally , intra - articular injections have been performed using anatomical landmarks to identify the correct trajectory for needle placement . however , different anatomical - guided injection techniques have yielded inconsistent intra - articular needle positioning due , in large part , to the fact that the physician can not directly visualize the area of interest , and variations in anatomy are common . 
 incorrect needle placement has been partially attributed to variable clinical outcomes.410 furthermore , inaccurate corticosteroid injections in the knee , for example , may result in post - injection pain , crystal synovitis , hemarthrosis , joint sepsis , and steroid articular cartilage atrophy , as well as systemic effects , such as fluid retention or exacerbation of hypertension or diabetes mellitus.1 therefore , identification of methods and proper training to aid in correct needle placement during these procedures is warranted . 
 various imaging modalities can be used to improve the accuracy of intra - articular injections , including fluoroscopy , computed tomography , and magnetic resonance imaging . 
 however , musculoskeletal ultrasound is one of the most practical because it is rapid , safe , relatively inexpensive , emits no ionizing radiation , and can be performed in the outpatient clinical setting.11,12 ultrasound utilizes high - frequency sound waves to visualize soft tissues and bony structures and is a f... \\ 
 \midrule 
Gold & \tiny intra - articular corticosteroid and hyaluronic acid injections provide short - term symptom amelioration for arthritic conditions involving structural damage or degenerative changes in the knee . 
 conventional palpation - guided anatomical injections frequently result in inaccurate needle placement into extra - articular tissue and adjacent structures . the purpose of this review was to determine the effect of ultrasound guidance on the accuracy of needle placement , clinical outcomes , and cost - effectiveness in comparison with anatomical landmark - guided intra - articular large joint injections , with particular emphasis on the knee . 
 a total of 13 relevant studies were identified ; five studied the knee , seven studied the shoulder , one used both the knee and shoulder , and none studied the hip . 
 ultrasound was used in seven studies ; the remaining studies utilized air arthrography , fluoroscopy , magnetic resonance arthrography , or magnetic resonance imaging . across all studies ( using all imaging modalities and all joints ) , needle placement accuracy ranged from 63\% to 100\% with ultrasound and from 39\% to 100\% with conventional anatomical guidance . 
 imaging guidance improved the accuracy of intra - articular injections of the knee ( 96.7\% versus 81.0\% , p \textless{} 0.001 ) and shoulder ( 97.3\% versus 65.4\% , p \textless{} 0.001 ) . 
 in particular , ultrasound guidance of knee injections resulted in better accuracy than anatomical guidance ( 95.8\% versus 77.8\% , p \textless{} 0.001 ) , yielding an odds ratio of 6.4 ( 95\% confidence interval 2.914 ) . 
 ultrasound guidance notably improves injection accuracy in the target intra - articular joint space of large joints including the knee . 
 the enhanced injection accuracy achieved with ultrasound needle guidance directly improves patient - reported clinical outcomes and cost - effectiveness . \\ 
 \midrule 
Model & \small objectivethe purpose of this review was to determine the effect of ultrasound guidance on the accuracy of needle placement, clinical outcomes, and cost - effectiveness in comparison with anatomical landmark - guided intra - articular large joint injections, with particular emphasis on the knee.methodsthe peer - reviewed literature was searched using medline and relevant bibliographies published in english language journals through december 31, 2011 that compared the accuracy of intra - articular injections with anatomical landmark - guided intra - articular large joint injections, with particular emphasis on the knee.resultsa total of 13 studies met the entry criteria. 
 needle placement accuracy ranged from 63\% to 100\% with ultrasound and from 39\% to 100\% with conventional anatomical guidance. 
 imaging guidance improved the accuracy of intra - articular injections of the knee ( 96.7\% versus 81.0\%, p 
 \textless{} 0.001 ) and shoulder ( 97.3\% versus 65.4\%, p 
 \textless{} 0.001 ). 
 imaging guidance improved the accuracy of intra - articular injections of the knee ( 96.7\% versus 81.0\%, p 
 \textless{} 0.001 ) and shoulder ( 97.3\% versus 65.4\%, p 
 \textless{} 0.001 ). 
 imaging guidance improved the accuracy of intra - articular injections of the knee ( 96.7\% versus 81.0\%, p 
 \textless{} 0.001 ) and shoulder ( 97.3\% versus 65.4\%, p 
 \textless{} 0.001 ). 
 in particular, ultrasound guidance of knee injections resulted in better accuracy than did anatomical guidance ( 95.8\% versus 77.8\%, p 
 \textless{} 0.001 ). 
 in particular, ultrasound guidance of knee injections resulted in better accuracy than did anatomical guidance ( 95.8\% versus 77.8\%, p 
 \textless{} 0.001 ). 
 in particular, ultrasound guidance of knee injections resulted in better accuracy than did anatomical guidance ( 95.8\% versus 77.8\%, p 
 \textless{} 0.001).conclusionthe use of imaging guidance, in particular ultrasound, improves the accuracy of intra - articular injections in large joints, including the knee. 
 furthermore, accurate ultrasound - guided intra - articular knee injections improve clinical outcomes and lower health care costs. \\ 
 \midrule 
Metrics & Rouge1: 62.21, Rouge2: 43.74, RougeL: 48.51, RougeLsum: 58.7, Summary length (tokens): 464
 \\ 

    \bottomrule
    \end{tabular}
}
\caption{Examples of the PubMed dataset using the model pubmed-4096-512 base diverse}
\label{tab:examples_pubmed-4096-512_base_diverse}
\end{table*}

\section{Data Details}
\label{sec:data_details}

\begin{figure*}[ht]
    \resizebox{\textwidth}{!}{
     \subfloat[
\textbf{Input Text}\\
Mean: 1289, Median: 1166\\
75-Quant: 1644, 95-Quant: 2290, Max: 3055\\
]
    {{
    \includegraphics[width=\textwidth/2]{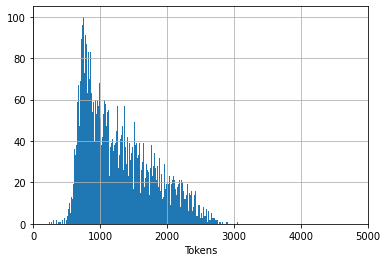} 
    }}
    \qquad
    \subfloat[
\textbf{Summary}\\
Mean: 179, Median: 157\\
75-Quant: 240, 95-Quant: 398, Max: 808\\
    ]
    {{
    \includegraphics[width=\textwidth/2]{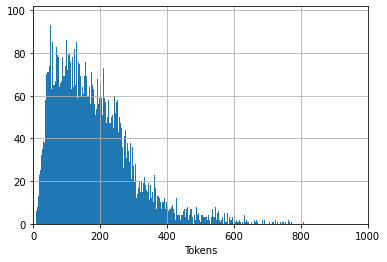} 
    }}
    }
    \caption{Histograms for the BillSum training set (18949 samples). }
    \label{fig:billsum_train}
\vspace{-5mm}
\end{figure*}

\begin{figure*}[ht]
    \resizebox{\textwidth}{!}{
    \subfloat[ 
\textbf{Input Text}\\
Mean: 1284, Median: 1164\\
75-Quant: 1629, 95-Quant: 2288, Max: 2957\\
]
    {{
    \includegraphics[width=\textwidth/2]{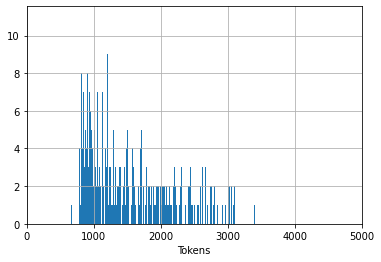} 
    }}
    \qquad
    \subfloat[
\textbf{Summary}\\
Mean: 179, Median: 156\\
75-Quant: 239, 95-Quant: 394, Max: 787\\
    ]
    {{
    \includegraphics[width=\textwidth/2]{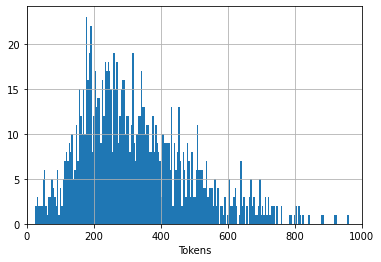} 
    }}
    }
    \caption{Histograms for the BillSum test set (3269 samples). }
    \label{fig:billsum_test}
\vspace{-5mm}
\end{figure*}

\begin{figure*}[ht]
    \resizebox{\textwidth}{!}{
    \subfloat[ 
\textbf{Input Text}\\
Mean: 3044, Median: 2572\\
75-Quant: 3996, 95-Quant: 7057, Max: 109759\\
]
    {{
    \includegraphics[width=\textwidth/2]{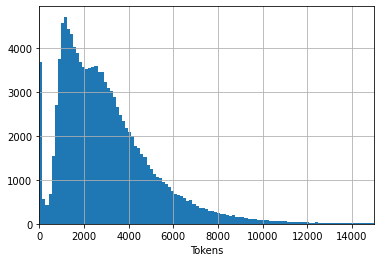} 
    }}
    \qquad
    \subfloat[
\textbf{Summary}\\
Mean: 202, Median: 208\\
75-Quant: 262, 95-Quant: 326, Max: 391\\
    ]
    {{
    \includegraphics[width=\textwidth/2]{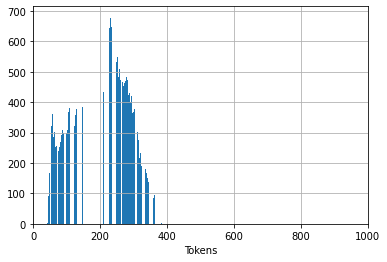} 
    }}
    }
    \caption{Histograms for the PubMed train set (119924 samples). }
    \label{fig:pubmed_train}
\vspace{-5mm}
\end{figure*}

\begin{figure*}[ht]
    \resizebox{\textwidth}{!}{
    \subfloat[ 
\textbf{Input Text}\\
Mean: 3112, Median: 2609\\
75-Quant: 4011, 95-Quant: 6968, Max: 119269\\
]
    {{
    \includegraphics[width=\textwidth/2]{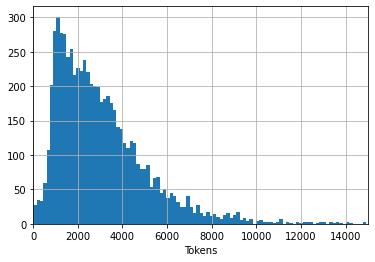} 
    }}
    \qquad
    \subfloat[
\textbf{Summary}\\
Mean: 203, Median: 209\\
75-Quant: 263, 95-Quant: 330, Max: 518\\
    ]
    {{
    \includegraphics[width=\textwidth/2]{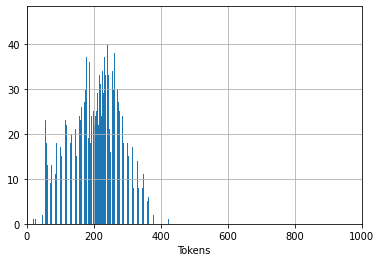} 
    }}
    }
    \caption{Histograms for the PubMed validation set (6633 samples). }
    \label{fig:pubmed_validation}
\vspace{-5mm}
\end{figure*}

\begin{figure*}[ht]
    \resizebox{\textwidth}{!}{
    \subfloat[ 
\textbf{Input Text}\\
Mean: 3093, Median: 2596\\
75-Quant: 3964, 95-Quant: 6985, Max: 48750\\
]
    {{
    \includegraphics[width=\textwidth/2]{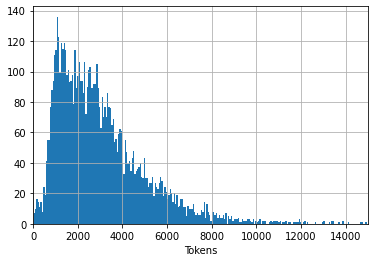} 
    }}
    \qquad
    \subfloat[
\textbf{Summary}\\
Mean: 205, Median: 213\\
75-Quant: 265, 95-Quant: 329, Max: 506\\
    ]
    {{
    \includegraphics[width=\textwidth/2]{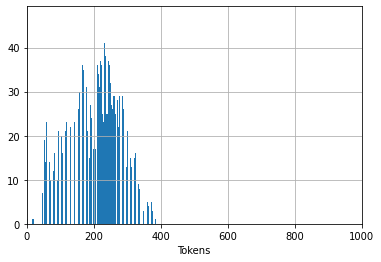} 
    }}
    }
    \caption{Histograms for the PubMed test set (6658 samples). }
    \label{fig:pubmed_test}
\vspace{-5mm}
\end{figure*}

We used our own tokenizer to calculate the number of tokens. In Tables \ref{fig:billsum_train}, and \ref{fig:billsum_test} we show the data length distributions for the BillSum train and test splits. In Tables \ref{fig:pubmed_train}, \ref{fig:pubmed_validation}, and \ref{fig:pubmed_test} we show the data length distributions for the PubMed train, validation and test splits.

\clearpage

\end{document}